\documentclass{article}

\usepackage{microtype}
\usepackage{graphicx}
\usepackage{wrapfig}
\usepackage{caption}
\usepackage{subfigure}
\usepackage{booktabs} 
\usepackage{pifont}

\usepackage{threeparttable}
\usepackage{amsmath}
\usepackage[table,xcdraw]{xcolor}
\usepackage{amssymb}
\usepackage{graphicx}
\usepackage{booktabs}
\definecolor{bg}{RGB}{176,226,255}
\definecolor{bonus_green}{RGB}{0,100,0}
\newcommand{\bbonus}[1]{{\textcolor{bonus_green}{$^{\uparrow#1}$}}}
\usepackage[colorlinks=true, allcolors=blue]{hyperref}
\definecolor{red}{HTML}{D9423C}
\definecolor{purple}{HTML}{854C98}

\usepackage{hyperref}

\usepackage{amsmath}
\usepackage{amssymb}
\usepackage{mathtools}
\usepackage{amsthm}

\usepackage[capitalize,noabbrev]{cleveref}

\theoremstyle{plain}
\newtheorem{theorem}{Theorem}[section]

\newtheorem{lemma}[theorem]{Lemma}

\theoremstyle{definition}
\newtheorem{definition}[theorem]{Definition}

\theoremstyle{remark}

\usepackage{PRIMEarxiv}

\usepackage[utf8]{inputenc} 
\usepackage[T1]{fontenc}    
\usepackage{hyperref}       
\usepackage{url}            
\usepackage{booktabs}       
\usepackage{amsfonts}       
\usepackage{nicefrac}       
\usepackage{microtype}      
\usepackage{lipsum}
\usepackage{fancyhdr}       
\usepackage{graphicx}       
\graphicspath{{media/}}     

\usepackage{multirow}
\usepackage[most]{tcolorbox}
\usepackage{float}
\definecolor{c1}{HTML}{F898CB}
\definecolor{c2}{HTML}{D9DEE7}
\definecolor{color_pl}{HTML}{7AC3DF}
\definecolor{sys_prompt}{HTML}{018B38}
\definecolor{question}{HTML}{652884}
\definecolor{CoT}{HTML}{F5A216}
\definecolor{answer}{HTML}{6A8EC9}

\title{Trust Region Preference Approximation: A simple and stable reinforcement learning algorithm for LLM reasoning}

\author{
Xuerui Su$^{1}$\thanks{~~Work done during internship at Microsoft Research AI for Science. Open-Source Research Project.} \quad \textbf{Shufang Xie}$^{2}$ \quad \textbf{Guoqing Liu}$^{2}$ \quad \textbf{
Yingce Xia}$^{2}$ \quad 
\textbf{Renqian Luo}$^{2}$ \\
\textbf{Peiran Jin}$^{2}$ \quad 
\textbf{Zhiming Ma}$^{4}$  \quad \textbf{Yue Wang}$^{3}$\footnotemark[2]  \quad
\textbf{Zun Wang}$^{2}$\footnotemark[2]\quad
\textbf{Yuting Liu}$^{1}$\thanks{~~Corresponding author.}  \quad
\\
$^1$Beijing Jiaotong University \quad
$^2$Microsoft Research AI for Science \quad
$^3$Zhongguancun Academy\\
$^4$University of Chinese Academy of Sciences, Academy of Mathemat-ics and Systems Science \vspace{2mm} \\
$^1$\texttt{\{24110486,ytliu\}@bjtu.edu.cn}\quad $^3$\texttt{yuewang\_yw@foxmail.com}\quad 
$^4$\texttt{mazm@amt.ac.cn} \\
$^2$\texttt{\{shufxi,guoqingliu,Yingce.Xia,renqianluo,peiranjin,zunwang\}@microsoft.com} 
}

\begin{document}
\maketitle

\begin{abstract}
Inference scaling further accelerates Large Language Models (LLMs) toward Artificial General Intelligence (AGI), with large-scale Reinforcement Learning (RL) to unleash long Chain-of-Thought reasoning. Reward-based optimization algorithms, such as Proximal Policy Optimization (PPO) and Group Relative Policy Optimization (GRPO) have achieved significant performance on reasoning tasks, whereas preference-based optimization algorithms such as Direct Preference Optimization (DPO) significantly improve the performance of LLMs on human alignment. However, despite the strong performance of reward-based optimization methods in alignment tasks , they remain vulnerable to reward hacking. Furthermore, preference-based algorithms (e.g. Online DPO) haven't yet matched the performance of reward-based optimization algorithms (like PPO) on reasoning tasks, making their exploration in this specific area still a worthwhile pursuit. Motivated by these challenges, we propose the Trust Region Preference Approximation (TRPA) algorithm, which integrates rule-based optimization with preference-based optimization for reasoning tasks. As a preference-based algorithm, TRPA naturally eliminates the reward hacking issue. TRPA constructs preference levels using predefined rules, forms corresponding preference pairs, and leverages a novel optimization algorithm for RL training with a theoretical monotonic improvement guarantee. Experimental results demonstrate that TRPA not only achieves competitive performance on reasoning tasks but also exhibits robust stability. The code of this paper are released and updating on \url{https://github.com/XueruiSu/Trust-Region-Preference-Approximation.git}.
\end{abstract}

\section{Introduction}

Inference scaling further accelerates Large Language Models (LLMs) \cite{ouyang2022training,grattafiori2024llama,claude3,DeepSeekV3} toward Artificial General Intelligence (AGI) \cite{goertzel2007artificial,wang2019defining,fei2022towards}. In this process, large-scale Reinforcement Learning (RL) techniques have played a crucial role in unleashing complex reasoning \cite{ouyang2022training,DPO,DeepSeekR1,jaech2024openai}. 
Works including OpenAI's O1 \cite{jaech2024openai}, DeepSeek R1 \cite{DeepSeekR1}, and Kimi k1.5 \cite{team2025kimi} have proposed employing reinforcement learning algorithms to boost reasoning performance, with DeepSeek R1 \cite{DeepSeekR1} and Kimi k1.5 \cite{team2025kimi} notably using rule-based rewards as learning signals for RL training.

Current large-scale RL algorithms can be broadly categorized into two classes: reward-based optimization and preference-based optimization. Reward-based methods, represented by PPO \cite{PPO-basedRLHF}, utilize carefully designed reward functions to drive the RL learning process, achieving notable success in human alignment tasks \cite{ouyang2022training}. Recent developments have seen the emergence of rule-based optimization algorithms, which define the value of behaviors through a set of predefined rules and subsequently use these rules to guide RL training. This approach has demonstrated significant advantages in improving safety-behavior accuracy \cite{mu2024rule}. Similar methodologies have been proposed in algorithms such as DeepSeek R1 \cite{DeepSeekR1}, Kimi k1.5 \cite{team2025kimi}, and Logic RL \cite{xie2025logic}, where a rule-designed reward function is employed to train the LLMs, resulting in substantial improvements in the reasoning capabilities of LLMs, as seen in representative methods like GRPO \cite{shao2024deepseekmath}. In contrast, preference-based optimization methods (as seen in methods like DPO \cite{DPO}, IPO \cite{IPO}, TDPO \cite{zeng2024token}, etc.) are predominantly applied to Human Alignment (HA) tasks and usually rely on a fixed offline dataset for preference optimization.

\begin{figure}[ht]
\vskip -0.1in
\begin{center}
\centerline{\includegraphics[width=0.85\columnwidth]{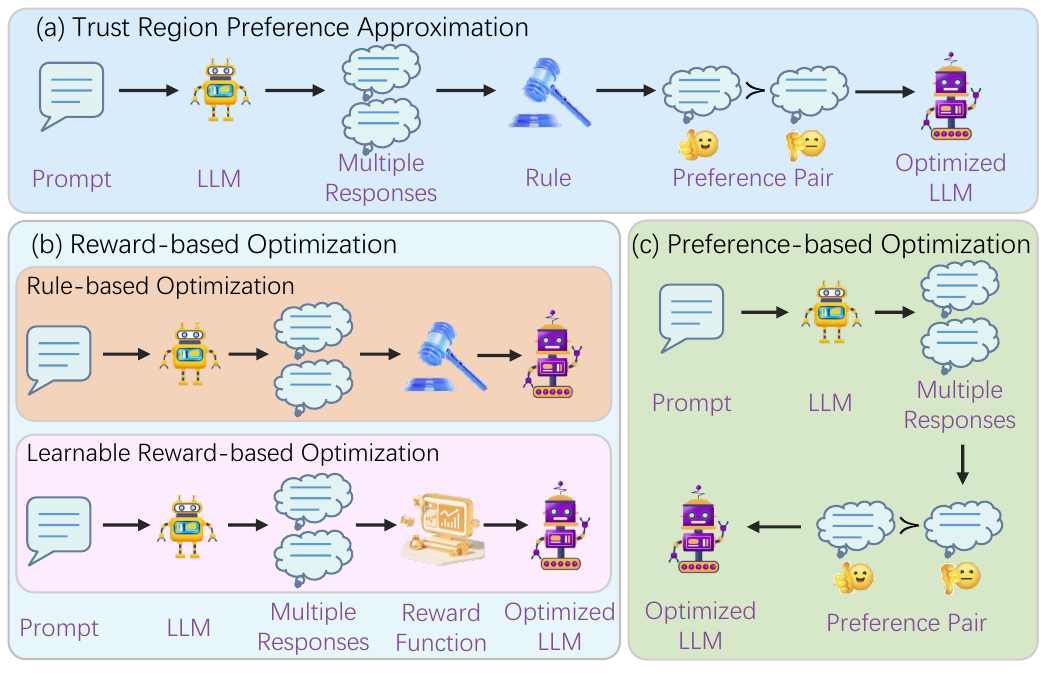}}
\caption{\centering (a) Trust Region Preference Approximation algorithm. (b) Reward-based Optimization algorithms with two reward scenarios, where Rule-based Optimization algorithms construct the reward function based on predefined rules (e.g., DeepSeek R1 \cite{DeepSeekR1}), and Learnable Reward-based Optimization algorithms implement training of a learnable reward model for subsequent RL training stages (e.g., InstructGPT \cite{ouyang2022training}). (c) Preference-based Optimization algorithms.}
\label{ROROPO_TRPA}
\end{center}
\vskip -0.2in
\end{figure}

Despite these advancements, current reward-based optimization algorithms face challenges related to reward hacking \cite{amodei2016concrete}, which will result in LLMs only receiving high-value rewards during RL training without genuinely learning or completing the intended task \cite{weng2024rewardhack,pan2022effects,di2022goal}. Moreover, preference-based optimization algorithms exemplified by DPO \cite{DPO} remain under-explored in LLM reasoning tasks. Recent studies indicates that comparing with PPO \cite{PPO-basedRLHF}, the Online DPO algorithm still achieves inferior performance \cite{zhang2025dpor1}. Furthermore, we reveal that the Online DPO algorithm exhibits a form of theoretical bias, which may explain its underperformance compared to algorithms such as GRPO \cite{shao2024deepseekmath} and PPO \cite{PPO-basedRLHF}. Thus, while preference-based optimization represents a novel RL paradigm, significant technical challenges remain in unlocking its full potential for LLM reasoning tasks. 

To overcome these challenges, this paper proposes a preference-based RL algorithm for LLM reasoning tasks named Trust Region Preference Approximation (TRPA) algorithm. TRPA leverages the technical route of rule-based and preference-based optimization by first designing preference levels based on predefined rules, then constructing preference pairs according to the corresponding levels, and finally employing the preference-based optimization algorithm for RL training with a theoretically monotonic improvement guarantee. See the detailed algorithm processes in Figure \ref{ROROPO_TRPA}. 


As a preference-based algorithm, TRPA naturally circumvents the complexities associated with designing reward functions, thereby reducing development overhead and eliminating reward hacking issues. In addition, we provide theoretical support to guarantee that TRPA achieves a Monotonic Improvement Guarantee toward the target distribution $\bar{\pi}^\tau(y|x)$ (see Eq.\ref{pi_bar}). 
Experimental results demonstrate that TRPA not only achieves compelling performance but also maintains excellent stability without encountering gradient explosion or other issues. Our contributions are as follows:
\begin{itemize}
    \item We propose the TRPA algorithm, which leverages preference-based optimization for simple and stable reinforcement learning to enhance the reasoning capabilities of LLMs.
    \item We offer a rigorous theoretical proof that guarantees TRPA's Monotonic Improvement Guarantee toward the target distribution $\bar{\pi}^\tau(y|x)$.
    \item We explore several key technologies  that further improve the performance of TRPA, resulting in competitive results on reasoning tasks.
\end{itemize}

\section{Related Work \& Preliminary}
In this section, we introduce related work closely connected to this paper, together with the notation used throughout the paper. Due to space limitations, see Appendix \ref{sec:method_discussion} for the discussion of the relationships and differences among Preference-based Optimization, Reward-based Optimization and Rule-based Optimization methods.

\ding{192} \textbf{Bradley-Terry (BT) model.} Denote reward function $r(\cdot,\cdot)$. The BT model \cite{bradley1952rank} $p^*$ is:
\begin{equation}
    \label{BT_model}
\begin{aligned}
p^*(z=1|y_1,y_2,x) \triangleq p^*(y_1\succ y_2|x) \triangleq \sigma(r(x, y_1)-r(x, y_2)),
\end{aligned}
\end{equation}
where $\sigma(\cdot)$ is the sigmoid function, event $y_1\succ y_2$ means response $y_1$ is preferred over $y_2$, $z\in\{0,1\}$ is the preference random variable and $p^*(z=0|y_1,y_2,x)\triangleq p^*(y_2\succ y_1|x)$. The first definition symbol establishes the relationship between the preference order $y_1\succ y_2$ and the random variable $z$, and the second definition symbol determines the mathematical form of the BT model. Unless otherwise specified, $ p^* $ in this paper is modeled by Eq.\ref{BT_model}. 

\ding{193} \textbf{Direct Preference Optimization (DPO).} The Direct Preference Optimization (DPO) \cite{DPO} leverages the optimal policy form of PPO based RLHF \cite{PPO-basedRLHF} as theoretical support by representing the comparison probability (calculated under the assumption of BT model) of human preferences through the ratio between the policy $\pi_\theta$ and the reference policy $\pi_{ref}$. This approach eliminates the need for explicitly modeling the reward function. Thus, DPO directly optimizes the policy by maximizing the log-likelihood function based on human preference feedback:
\begin{equation}\label{DPO_eq}
{\begin{gathered}
\mathcal{L}_{\mathrm{DPO}}\left(\pi_\theta ; \pi_{\text {ref }}\right)=-\mathbb{E}_{\left(x, y_1, y_2\right) \sim \mathcal{D}_R}\left[\log \sigma\left(\bar{h}_\theta\left(x, y_1, y_2\right)\right)\right],
\end{gathered}}
\end{equation}
where $\bar{h}_\theta\left(x, y_1, y_2\right)=\beta \log \frac{\pi_\theta\left(y_1 \mid x\right)}{\pi_{\text {ref }}\left(y_1 \mid x\right)}-\beta \log \frac{\pi_\theta\left(y_2 \mid x\right)}{\pi_{\text {ref }}\left(y_2 \mid x\right)}$, $\mathcal{D}_{R}\triangleq \{(x,y_1,y_2)|x\sim\mathcal{D},y_1,y_2\sim\pi_0(\cdot|x),(y_1\succ y_2)\sim p^*(z|y_1,y_2,x)\}$, $\pi_0$ is an offline data sampling policy and $\mathcal{D}$ is a given prompt distribution. 

\ding{194} \textbf{Online Direct Preference Optimization (Online DPO).} As a trivial solution to design DPO as an online algorithm, Online DPO is a variant of Online AI Feedback (OAIF) for Direct Alignment from Preference (DAP) methods when the loss function is the DPO loss \cite{guo2024direct}. In Online DPO, the response pairs used to construct the preference pairs come from the current LLM $\pi_t(\cdot|x)$ to be optimized:
\begin{equation}
\small{\begin{gathered}
\mathcal{L}_{\mathrm{Online\ DPO}}\left(\pi_{\theta} ; \pi_{\text {ref }}\right)=-\mathbb{E}_{\left(x, y_1, y_2\right) \sim \mathcal{D}_\theta}\left[\log \sigma\left(\bar{h}_{\theta}\left(x, y_1, y_2\right)\right)\right],
\end{gathered}}
\end{equation}
where $\mathcal{D}_{\theta}\triangleq \{(x,y_1,y_2)|x\sim\mathcal{D},y_1,y_2\sim\pi_{\theta}(\cdot|x),(y_1\succ y_2)\sim p^*(1|y_1,y_2,x)\}$.

\ding{195} \textbf{Preference Approximation (PA) algorithm.} To ensure the consistency between the algorithm's name and its meaning, as well as to avoid potential misinterpretations, this paper renames the "Preference Reward Approximation-Posterior (PRA-P) method" in the framework named UDRRA \cite{su2025reveal} to the "Preference Approximation (PA) algorithm". 
Similar with Online DPO, the PA algorithm also deploys training of LLMs on $\mathcal{D}_{\theta}$. Differently, the PA algorithm uses the following loss function:
\begin{equation}
    \mathcal{L}_{\mathrm{PA}}(\pi_\theta; \pi_{\text {ref }})
=-\mathbb{E}_{(x,y_1,y_2)\sim \mathcal{D}_{\theta}}\left[\log \sigma\left(\bar{h}_\theta\left(x, y_1, y_2\right)\right)\right]+\mathbb{E}_{x\sim \mathcal{D},y_1,y_2\sim \pi_\theta(y|x)}\left[M(x,y_1,y_2)\right],
\end{equation}
where $M(x,y_1,y_2)=\sum_{z=0,1}p^*(z|y_1,y_2,x)\log p^*(z|y_1,y_2,x)$. In section \ref{PBA_sec}, we will demonstrate that the PA algorithm and the PPO algorithm share the same target distribution. In contrast, Online DPO lacks such theoretical support, despite the fact that DPO is derived from PPO.

\section{Method}
We propose the Trust Region Preference Approximation (TRPA) algorithm for stable RL training, which leverages the technical routes of rule-based and preference-based optimization.

\subsection{Trust Region Preference Approximation (TRPA)}
Trust Region Preference Approximation (TRPA) is a preference-based optimization method based on the rule-based optimization scheme. See Figure \ref{ROROPO_TRPA} for the training process of TRPA. The TRPA algorithm utilizes a predefined rule to classify the multiple responses generated by LLM into different preference levels. The preference level represents the degree to which the corresponding response is preferred based on our predefined rule. Then we construct preference pairs between responses which have the same prompt but different preference levels. Finally, these preference pairs are used for optimization under the loss function of the TRPA algorithm. Given the reference policy $\pi_{ref}(y|x)$ and the policy $\pi_{\theta_{\text{old}}}$ which is the policy before the current optimization step and used for generating data $\mathcal{D}_{\text{old}}$, the TRPA algorithm trains the policy model $\pi_{\theta}$ under the following loss function:
\begin{equation}\label{TRPRA_eq}
\small{\begin{aligned}
\mathcal{L}_{\text{TRPA}}\left(\pi_{\theta};\pi_{\text {ref }}\right)=-&\mathbb{E}_{\left(x, y_1, y_2\right) \sim \mathcal{D}_{\text{old}}}\left[\log \sigma\left(\beta(y_1) \log \frac{\pi_\theta\left(y_1 \mid x\right)}{\pi_{\text {ref }}\left(y_1 \mid x\right)}-\beta(y_1) \log \frac{\pi_\theta\left(y_2 \mid x\right)}{\pi_{\text {ref }}\left(y_2 \mid x\right)}\right)\right]\\
+&\lambda\mathbb{E}_{x\sim \mathcal{D}}\left[{\mathrm{D}_{\mathrm{KL}}(\pi_{\theta_{\text{old}}}(\cdot|x)||\pi_{\theta}(\cdot|x))}\right],
\end{aligned}}
\end{equation}
where  $\mathcal{D}_{\text{old}}\triangleq \{(x,y_1,y_2)|x\sim\mathcal{D},y_1,y_2\sim\pi_{\theta_{\text{old}}}(\cdot|x),(y_1\succ y_2)\sim p^*(1|y_1,y_2,x)\}$, $\lambda$ is a hyper-parameter and $\beta(y_1)$ is the KTPO coefficient which will be discussed in Section \ref{TRPA_details}.\ding{194}.

\subsection{Technical Details}\label{TRPA_details}
Building on the principles of preference-based optimization, we develop several key techniques tailored to the TRPA framework to support more effective reasoning and planning in LLMs.

\ding{192} \textbf{KL Regularization.} Maintaining controlled update steps facilitates smoother convergence during training. TRPA incorporates the constraint 
${\mathrm{D}_{\mathrm{KL}}(\pi_{\theta_{\text{old}}}(\cdot|x)||\pi_{\theta}(\cdot|x))}$ to effectively leverage the benefits of trust region principle. By limiting the KL divergence between successive policies, TRPA ensures gradual updates, thereby preventing abrupt changes that could destabilize the training process.

\ding{193} \textbf{Definition of Rule and Preference Levels.} We adopt the same reward categorization rule as the rule-based optimization algorithms, such as Logic RL \cite{xie2025logic} and DeepScaleR \cite{deepscaler2025}. Specifically, we define four preference levels to evaluate LLM-generated responses, ranging from Level 4 (lowest) to Level 1 (highest), based on both formatting and factual correctness. 
\begin{tcolorbox}[
    colframe=color_pl, 
    colback=blue!5!white, 
    coltitle=white, 
    fonttitle=\bfseries, 
    title=Preference Levels\label{Preference_Levels}, 
    arc=3mm, 
    boxrule=0.5mm, 
]
$\bullet$ Preference Level 4: Responses that violate basic formatting requirements. Specifically, we require responses to follow the standard Chain-of-Thought (CoT) format: the reasoning process and final answer must be enclosed in \texttt{<think>...</think>} and \texttt{<answer>...</answer>} tags, respectively. For math-related problems, the final answer should be wrapped in \texttt{\textbackslash\textbackslash boxed\{\}}.\\
$\bullet$ Preference Level 3: Responses that meet formatting requirements but exhibit incomplete or illogical reasoning, indicating a partial understanding of the expected format. \\ 
$\bullet$ Preference Level 2: Responses that are correctly formatted but still contain factual error. \\
$\bullet$ Preference Level 1: Responses that are both well-formatted and factually correct.
\end{tcolorbox}
Here we explain why we choose the BT model \cite{bradley1952rank} instead of listwise ranking models (e.g. \cite{luce1959individual}). In our experiments, we observe that rollouts under the same prompt typically yield outputs falling into two adjacent preference levels, supporting the assumption of binary comparisons of the BT model.



\ding{194} \textbf{Kahneman-Tversky Preference Optimization (KTPO).} To better guide the optimization process, we introduce the KTPO technique, inspired by Prospect Theory (Kahneman–Tversky model) \cite{altman2010prospect}, which suggests that humans exhibit asymmetric sensitivity to gains and losses. Motivated by this asymmetry, we design the hyperparameter $\beta(y)$ to be anisotropic, assigning larger values when $y$ corresponds to a response of higher preference level (e.g., Level 1). Let $\beta$ be the base hyperparameter and $N$ be the Kahneman–Tversky factor, then
\begin{equation}
    \beta(y) = 
\begin{cases} 
N\cdot\beta, & \text{if } y \text{ belongs to Preference Level 1}, \\
\beta, & \text{otherwise}.
\end{cases}
\end{equation}
TRPA introduces the adaptive hyperparameter $\beta$ (i.e. KTPO) for the first time in the LLMs Reasoning tasks. Experimental results show that this method can significantly improve model performance. It is worth noting that in the Human Alignment tasks, there are a series of preference optimization algorithms (such as WPO \cite{zhou2024wpo}, Adaptive DPO \cite{hong2024adaptive}, NS-DPO \cite{son2024right}, $\beta$-DPO \cite{wu2024beta}, KTO \cite{KTO}, etc.) that follow the idea of anisotropy to improve the algorithm performance through the design of $\beta$. Given the focus of this study, we did not conduct a detailed horizontal comparison of existing methods, but their ideas have certain commonalities with this study.

\ding{195} \textbf{Prompt-wised Optimization Strategy.} In both the early and late stages of optimization, LLMs often generate responses with identical preference levels (or function values) under the same prompt. This typically arises from undertraining in the early phase and overtraining or convergence in the later phase. For example, in GRPO \cite{shao2024deepseekmath}, this phenomenon leads to normalized rewards being zero, which in turn causes the gradient of the loss function to vanish, rendering the optimization ineffective. In contrast, the TRPA algorithm naturally accommodates such scenarios through a Prompt-wised Optimization Strategy: when all responses have a preference level of 1, they are treated as $y_1$ and the corresponding $y_2$ terms are omitted during optimization; conversely, when all responses share a non-1 but identical preference level, they are treated as $y_2$, and the $y_1$ terms are excluded accordingly.

\section{Theorem}
In this section, we will perform a theoretical analysis of why TRPA ensures a Monotonic Improvement Guarantee towards the target distribution $\bar{\pi}^\tau(y|x)$. And in contrast, Online DPO \cite{guo2024direct} does not have this guarantee and does not take $\bar{\pi}^\tau(y|x)$ as the target distribution, even though this algorithm has a very similar loss function to TRPA. In Section \ref{PBA_sec}, we introduce the concept of the Posterior Boltzmann Approximation (PBA) algorithm, which takes $\bar{\pi}^\tau(y|x)$ as the target distribution, and prove the Online DPO algorithm is not a PBA algorithm, but the PPO and PA algorithms are.
In section \ref{theorem_subsec}, we demonstrate that compared to the PA algorithm, which is a PBA algorithm but has a loss function that is difficult to estimate, the TRPA algorithm has both the monotonic improvement guarantee of the PA algorithm and the loss function of the TRPA algorithm is easier to estimate.
\subsection{Posterior Boltzmann Approximation Algorithm}\label{PBA_sec}
\begin{definition}\label{PBA_def}
An algorithm qualifies as a Posterior Boltzmann Approximation (PBA) algorithm if and only if its loss function employs $\bar{\pi}^\tau(y|x)$ as the target distribution:
\begin{equation}\label{pi_bar}
    \small{\bar{\pi}^\tau(y|x)=\frac{1}{Z(x)} \pi_{\mathrm{ref}}(y \mid x) \exp \left(\frac{1}{\beta} r(x, y)\right),\quad Z(x)=\sum_y \pi_{\mathrm{ref}}(y \mid x) \exp \left(\frac{1}{\beta} r(x, y)\right).}
\end{equation} 
\end{definition}
$\bar{\pi}^\tau(y|x)$ is exactly the optimal solution to the KL-constrained reward maximization objective of PPO\cite{PPO-basedRLHF} and the DPO objective described in \cite{DPO}. However, we prove that the trivial online extension of DPO, namely Online DPO, does not satisfy the definition of a PBA method, indicating that designing an online algorithm for DPO is a non-trivial task. In contrast, we propose that the PA algorithm is a more suitable solution to construct online preference optimization algorithms. 



\begin{figure}[ht]
\begin{center}
\centerline{\includegraphics[width=0.95\columnwidth]{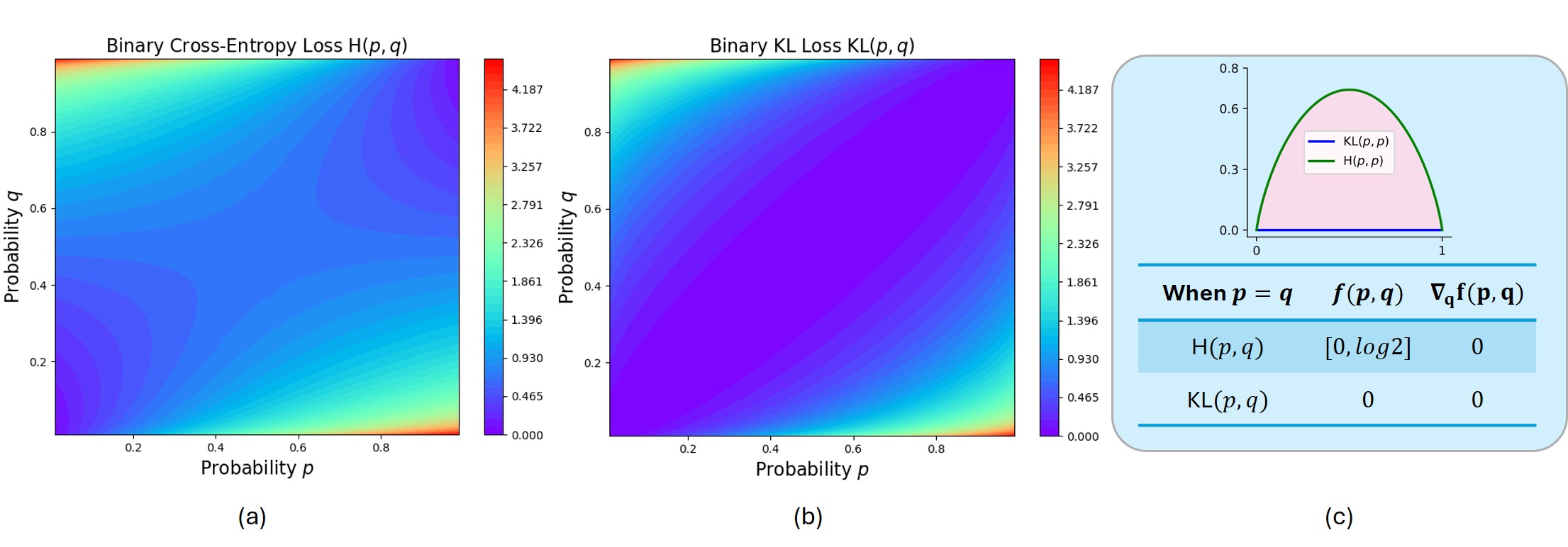}}
\caption{\centering (a) Contour map of the cross-entropy loss function. (b) Contour map of the KL loss function. (c) Comparison of the KL loss and cross-entropy loss when $p=q$.}
\label{hpq_kl_figure}
\end{center}
\vskip -0.2in
\end{figure}

\begin{lemma}\label{Online_DPO_is_not_PBA}
The Online DPO algorithm is not a PBA algorithm.
\end{lemma}
\begin{lemma}\label{TRPA_is_PBA}
The PA algorithm is a PBA algorithm.
\end{lemma}
Lemma \ref{Online_DPO_is_not_PBA} and Lemma \ref{TRPA_is_PBA} demonstrate that the Online DPO and PA algorithms have completely opposite conclusions (see proof on Appendix \ref{Online_DPO_is_not_PBA_proof}, \ref{TRPA_is_PBA_proof}). This phenomenon stems from the difference between the loss function landscape of cross-entropy and KL divergence under the setting of online learning. With straightforward algebraic derivation, we have that the correspondence of the Online DPO and PA algorithms with cross-entropy and KL divergence loss functions, as expressed in Eq.\ref{OnlineDPO} and Eq.\ref{PA_KL_eq}. The optimal solution set for cross-entropy consists of the points $(0,0)$ and $(1,1)$, whereas for KL divergence, the optimal solutions form the set of all points on the line $ y = x $ within the range $[0,1] \times [0,1]$. We show the Loss function landscape of cross-entropy and KL in Figure \ref{hpq_kl_figure}.
\begin{equation}\label{OnlineDPO}
    \begin{aligned}
&\mathcal{L}_{\mathrm{Online\ DPO}}\left(\pi_{\theta} ; \pi_{\text {ref }}\right)=\mathbb{E}_{x\sim \mathcal{D},y_1,y_2\sim \pi_\theta(y|x)}\left[\mathrm{H}(p^*(z|y_1,y_2,x)||p_\theta(z|y_1,y_2,x))\right],
\end{aligned}
\end{equation}
where $\mathrm{H}(p||q)=-\mathrm{E}_p[\log q]$ is the cross-entropy of the distribution $q$ relative to a distribution $p$, $p_\theta(z=1|y_1,y_2,x)=\sigma\left(\bar{h}_\theta(x,y_1,y_2)\right)$ and $p_\theta(z=0|y_1,y_2,x)=\sigma\left(\bar{h}_\theta(x,y_2,y_1)\right)$.
\begin{equation}\label{PA_KL_eq}
\begin{aligned}
    \mathcal{L}_{\mathrm{PA}}(\pi_\theta; \pi_{\text {ref }})
=\mathbb{E}_{x\sim \mathcal{D},y_1,y_2\sim \pi_\theta(y|x)}\left[\mathrm{D}_{\mathrm{KL}}(p^*(z|y_1,y_2,x)||p_\theta(z|y_1,y_2,x))\right],
\end{aligned}
\end{equation}
where $\mathrm{D}_{\mathrm{KL}}(p||q)=\mathrm{E}_p[\log p-\log q]$ is the KL divergence.

When we require the gradient of the loss function to be 0, for online learning, the zero-order and first-order derivatives (the underlined parts) of the integrated distribution $f(x,y_1,y_2;\pi_\theta)$ are required to be 0 at the optimal point. This requirement is inconsistent with supervised learning; supervised learning usually only requires the zero-order optimality of the integrated distribution to be 0.
\begin{equation}\label{online_learning_eq}
\small{\begin{aligned}
    &\quad\nabla_\theta\mathcal{L}_{\mathrm{online\ learning}}(\pi_\theta)=\nabla_\theta\mathbb{E}_{x\sim \mathcal{D},y_1,y_2\sim \pi_\theta(y|x)}\left[{f(x,y_1,y_2;\pi_\theta)}\right]\\
    &\small{=\sum_{x,y_1,y_2} \mathcal{D}(x) \left(\underline{f(x,y_1,y_2;\pi_\theta)}\nabla_\theta (\pi_\theta(y_1|x)\pi_\theta(y_2|x))+ \pi_\theta(y_1|x)\pi_\theta(y_2|x)\underline{\nabla_\theta f(x,y_1,y_2;\pi_\theta)}\right)},
\end{aligned}}
\end{equation}
where $\mathcal{D}(x)$ is an arbitrary distribution,  $f(x,y_1,y_2;\pi_\theta)$ can be $\mathrm{H}(p^*(z|y_1,y_2,x)||p_\theta(z|y_1,y_2,x))$ or $\mathrm{D}_{\mathrm{KL}}(p^*(z|y_1,y_2,x)||p_\theta(z|y_1,y_2,x))$. Based on Eq.\ref{online_learning_eq}, we can know that $\nabla_\theta\mathcal{L}_{\mathrm{online\ learning}}(\pi_\theta)=0$ if and only if $f(x,y_1,y_2;\pi_\theta)=0$ and $\nabla_\theta f(x,y_1,y_2;\pi_\theta)=0$.

For preference optimization, the optimal solution is $p^*(z|y_1,y_2,x)=p_\theta(z|y_1,y_2,x)$, that is, the fitted preference distribution $p_\theta$ is consistent with the ground truth preference distribution $p^*$. For cross-entropy $\mathrm{H}(\cdot||\cdot)$, the zero-order derivative isn't 0 when $p^*(z|y_1,y_2,x)=p_\theta(z|y_1,y_2,x)$ unless $p^*(z|y_1,y_2,x), p_\theta(z|y_1,y_2,x)$ are equal to 0 or 1. This will cause the gradient of the entire loss function to be non-zero in the ideal case ($p^*(z|y_1,y_2,x)=p_\theta(z|y_1,y_2,x)$), resulting in unpredictable optimization results. However, the KL divergence doesn't have this issue.

\subsection{Monotonic Improvement Guarantee of TRPA for the PA algorithm}\label{theorem_subsec}
Although as a PBA algorithm, the PA algorithm has good properties on the target distribution, the PA algorithm still faces a challenge: the $M(x,y_1,y_2)$ term is hard to calculate. $M(x,y_1,y_2)=\sum_{z=0,1}p^*(z|y_1,y_2,x)\log p^*(z|y_1,y_2,x)$ is constructed by the ground truth preference distribution $p^*$ which is usually implicit in the preference data. In practice, $M(x,y_1,y_2)$ is usually difficult to estimate accurately. Therefore, in this section, we propose to use the TRPA algorithm which has a simpler loss function as a practical algorithm from these theoretical foundations. At the same time, we provide a monotonic improvement guarantee for the PA algorithm to ensure the PBA algorithm properties of the TRPA algorithm by Theorem \ref{TRPA}. See Appendix \ref{TRPRA_bound_proof} for proof.
\begin{theorem}\label{TRPA}
Denote 
\begin{equation}
    \begin{aligned}
    \bar{L}^{\pi_{\theta_{\text{old}}}}(\pi_{\theta_{\text{new}}})=&-\mathbb{E}_{(x,y_1,y_2)\sim \mathcal{D}_{{\text{old}}}}\left[\log\sigma\left( \bar{h}_{\theta_{\text{new}}}\left(x, y_1, y_2\right)\right)\right]+\mathbb{E}_{x\sim \mathcal{D},y_1,y_2\sim \pi_{\theta_{\text{old}}}(\cdot|x)}\left[M(x,y_1,y_2)\right].
\end{aligned}
\end{equation}
Then we have the following inequality hold:
\begin{equation}\label{TRPRA_bound_eq_2}
    \small{
    \begin{aligned}
        \bar{L}^{\pi_{\theta_{\text{new}}}}(\pi_{\theta_{\text{new}}}) \leq &\bar{L}^{\pi_{\theta_{\text{old}}}}(\pi_{\theta_{\text{new}}}) + \lambda\sqrt{D_{\mathrm{KL}}^{\max }(\pi_{\theta_{\text{old}}}|| \pi_{\theta_{\text{new}}})},
    \end{aligned}
    }
\end{equation}
where $\lambda=4(\bar{U}_r + 2\log2)$ and $\bar{U}_r=\max_{x,y_1,y_2}  \mathrm{D}_{\mathrm{KL}}(p^*(z|y_1,y_2,x)||\bar{p}_{\theta_{\text{new}}}(z|x,y_1,y_2))$.
\end{theorem}
We briefly describe the relationship between the TRPA algorithm and Theorem \ref{TRPA}.
First, the $\mathbb{E}_{x\sim \mathcal{D},y_1,y_2\sim \pi_{\theta_{\text{old}}}(y|x)}\left[M(x,y_1,y_2)\right]$ term of $L^{\pi_{\theta_{\text{old}}}}(\pi_{\theta_{\text{new}}})$ is omitted as it is irrelevant to $\pi_{\theta_{\text{new}}}$. Second, when $D_{\mathrm{KL}}^{\max }(\pi_{\theta_{\text{old}}}|| \pi_{\theta_{\text{new}}})\rightarrow 0$, the $\sqrt{\mathrm{D}_{\text{KL}}(\pi_{\theta_{\text{old}}}||\pi_{\theta_{\text{new}}})}$ term's smooth coefficient will tend to $\infty$. Therefore, we use $D_{\mathrm{KL}}^{\max }(\pi_{\theta_{\text{old}}}|| \pi_{\theta_{\text{new}}})$ as the trust region regularization item of TRPA. Finally, we get the final loss function of Eq.\ref{TRPRA_eq}. The two changes are guaranteed by Theorem \ref{TRPA} to ensure the monotonic improvement guarantee of TRPA about $\bar{L}^{\pi_{\theta_{\text{new}}}}(\pi_{\theta_{\text{new}}})$.

\section{Experiment}
Here, we report our experiment results of TRPA on the K\&K logic puzzle dataset \cite{xie2024memorization} and Math dataset \cite{deepscaler2025}. See Table \ref{logic_table} and Table \ref{math_table} for main results. We also provide a case study in Appendix \ref{Case_Study}.
\subsection{Main Results}\label{sec:experiment_main_results}
For logic puzzle task, we used 4.5k data with a range of 3 to 7 number of people for TRPA training, using \texttt{Qwen2.5-7B-Instruct-1M} as the base model, with a batch size of 4, rollout time of 8, and trained on 4$\times$A100-80GB for 7,200 steps with a constant learning rate of $4\times10^{-7}$, max response length of 4,096, and temperature parameter of 1. We kept the data size (batch\_size$\times$rollout\_times$\times$training\_steps) consistent with Logic-RL \cite{xie2025logic}.

\begin{table}[h]
    \centering
    \renewcommand{\arraystretch}{1.3}
    \Large
    \vspace{-0.1in}
    \caption{\centering Evaluation of reasoning and general models on the K\&K logic puzzle dataset across different difficulty levels (pass@1).}
    \resizebox{0.85\linewidth}{!}{
    \begin{tabular}{@{}l *{5}{c}|*{2}{c}|*{1}{c} @{}}
    \toprule
    \multirow{3}{*}{\centering\textbf{Model}} & \multicolumn{6}{c}{\textbf{\quad \quad \quad \quad \quad Difficulty by Number of People}} \\
    \cmidrule(r){2-8}
     & \multicolumn{1}{c}{3} & \multicolumn{1}{c}{4} & \multicolumn{1}{c}{5} & \multicolumn{1}{c}{6} & \multicolumn{1}{c}{7} & \multicolumn{1}{c}{\textbf{2} (OOD)} & \multicolumn{1}{c}{\textbf{8} (OOD)} & \multicolumn{1}{c}{\multirow{2}{*}[+2.5ex]{Avg.}} \\
    \midrule
    \textbf{o3-mini-high} & 0.98 & 0.97 & \textbf{0.95} & \textbf{0.94} & 0.89 & \textbf{0.99} & 0.83 & 
 {0.935} \\
    \textbf{o1-2024-12-17} & 0.51 & 0.38 & 0.38 & 0.35 & 0.30 & 0.83 & 0.20 & 0.421\\
    \textbf{Deepseek-R1} & 0.73 & 0.77 & 0.78 & 0.75 & 0.88 & 0.91 & 0.83 & 0.807\\
    \midrule
    \textbf{GPT-4o} & 0.57 & 0.49 & 0.32 & 0.23 & 0.21 & 0.68 & 0.11 & 0.373\\
    \textbf{GPT-4o-mini} & 0.42 & 0.34 & 0.17 & 0.09 & 0.10 & 0.63 & 0.01 & 0.251\\
    \textbf{NuminaMath-7B-CoT} & 0.13 & 0.12 & 0.05 & 0.01 & 0.00 & 0.28 & 0.00 & 0.084 \\
    \textbf{Deepseek-Math-7B} & 0.21 & 0.08 & 0.06 & 0.02 & 0.00 & 0.35 & 0.00 & 0.103\\
    \textbf{Qwen2.5-Base-7B} & 0.34 & 0.16 & 0.09 & 0.00 & 0.00 & 0.41 & 0.00 & 0.143 \\
    \textbf{Logic-RL} & \textbf{0.99} & {0.94} & 0.92 & 0.91 & 0.80 & \textbf{0.99} & 0.67 & 0.889 \\
    \midrule
    \textbf{Qwen2.5-7B-Instruct-1M} & 0.40 & 0.25 & 0.11 & 0.06 & 0.02 & 0.49 & 0.01 & 0.191 \\
    \rowcolor{bg!70}
    \textbf{+TRPA (Ours)} & \textbf{0.99} & \textbf{0.98} & \textbf{0.95} & {0.92} & \textbf{0.91} & 0.96 & \textbf{0.86} & \textbf{0.938}\\
    \bottomrule
    \end{tabular}
    }
    \label{logic_table}
\end{table}

Table \ref{logic_table} shows the performance comparison of the LLM trained by our TRPA algorithm compared with other representative LLMs. The average performance of TRPA reached SOTA 94\%. In the in-distribution test results, the SOTA effect was achieved in most cases (except number of people = 6). Moreover, in the out-of-distribution test results, we found that TRPA significantly improved the model's ability to generalize to more difficult problems. During the entire training process, LLM was never trained on data with a number of people of 8. But in the case of number of people = 8, TRPA achieved an accuracy of 86\% that surpassed Deepseek-R1 and o3-mini-high, which fully demonstrates that TRPA enhances the knowledge generalization performance of the LLM.

For the math reasoning tasks (Table \ref{math_table}), we use the DeepScaleR-Preview-Dataset \cite{deepscaler2025} as the training set and evaluate on AIME 2024, MATH 500, AMC 2023, Minerva Math, and Olympiad Bench. The DeepScaleR-Preview-Dataset comprises AIME (1984–2023), AMC (excluding 2023), Omni-MATH \cite{gao2024omni}, and Still \cite{min2024imitate} datasets, totaling approximately 40.3k samples. Due to computational constraints, we conduct training on 4$\times$A100-80GB GPUs, with a fixed learning rate of $4\times10^{-7}$, using \texttt{DeepSeek-R1-Distill-Qwen-7B (8k)} (trained with a maximum response length of 8k) as the base model, a batch size of 4, rolled out times of 8, maximum response length of 8,192, and temperature of 1. After training for approximately 0.8 epochs (8k steps), the model's performance on AIME 2024 matches that of \texttt{DeepSeek-R1-Distill-Qwen-7B (32k)} distilled under the maximum response length of 32k. Compared to \texttt{DeepSeek-R1-Distill-Qwen-7B (8k)}, our model achieves an average performance improvement of 10.5\%, demonstrating the effectiveness of TRPA in training on complex mathematical datasets.  
\begin{table}[h]
    \centering
    \renewcommand{\arraystretch}{1.3}
    \Large
    \vspace{-0.1in}
    \caption{Evaluation of reasoning and general models on the Math (pass@1).}
    \resizebox{0.9\linewidth}{!}{
    \begin{tabular}{@{}l *{5}{c}|*{1}{c} @{}}
    \toprule
     {\centering\textbf{Model}} & AIME 2024 &  MATH 500 & AMC 2023 & Minerva Math & Olympiad Bench & Avg. \\
    \midrule
    \textbf{O1-Preview}	&0.40	&0.81	&-&	-&	-&	-\\
    \textbf{GPT-4o-2024-08-06}	&-	&0.81	&-&	0.37&	0.43&	-\\
    \textbf{Deepseek-R1} & 0.80 & 0.97 &-&	-&	-&	-  \\
    \midrule
    \textbf{Qwen2.5-Math-7B-Instruct} & 0.13	&0.80&	0.51 &	0.35&	0.41&	0.44\\
    \textbf{rStar-Math-7B}&	0.27& 0.78& 0.48&	-&	0.47&	- \\
    \textbf{Eurus-2-7B-PRIME}&	0.28 &	0.79&	0.58	& 0.39	&0.42&	0.49\\
    \textbf{Qwen2.5-7B-SimpleRL}&	0.27&	0.82&	0.63&	0.40 &	0.43	&0.51\\
    \textbf{DeepSeek-R1-Distill-Qwen-1.5B}&	0.29&	0.83&	0.63 &	0.27&	0.43	&0.49\\
    \textbf{DeepScaleR-1.5B-Preview}	&0.43	&0.88	&0.74&	0.30&	0.50&	0.57\\
    \textbf{DeepSeek-R1-Distill-Qwen-7B (32k)} & 0.56 & 0.93 &-&	-&	-&	-  \\
    \midrule
    \textbf{DeepSeek-R1-Distill-Qwen-7B (8k)} & 0.43   & 0.86  & 0.70 & 0.35 & 0.47 & 0.57 \\
    \rowcolor{bg!70}
    \textbf{+TRPA (Ours)} & {0.57} & {0.87} & \textbf{0.78} & {0.36} & \textbf{0.55} & 0.63\bbonus{10.5\%} \\
    \bottomrule
    \end{tabular}
    }
    \vspace{-0.1in}
    \label{math_table}
\end{table}
    


\subsection{Ablation Study}
In the following content, we analyze the effectiveness of the TRPA algorithm and its related techniques through a series of ablation experiments. We mainly focus on the following questions: \ding{192} Q1: How does TRPA perform differently compared to other works? \ding{193} Q2: What benefits does the KL divergence used by TRPA bring compared to Online DPO? \ding{194} Q3: The trends of Response Length and Entropy as RL training progresses. 
\ding{195} Q4: The role of KTPO technique. \ding{196} Q5: Does TRPA have the simultaneous change problem that preference optimization usually faces?

\begin{figure}[ht]
\vspace{-0.05in}
\begin{center}
\centerline{\includegraphics[width=0.9\columnwidth]{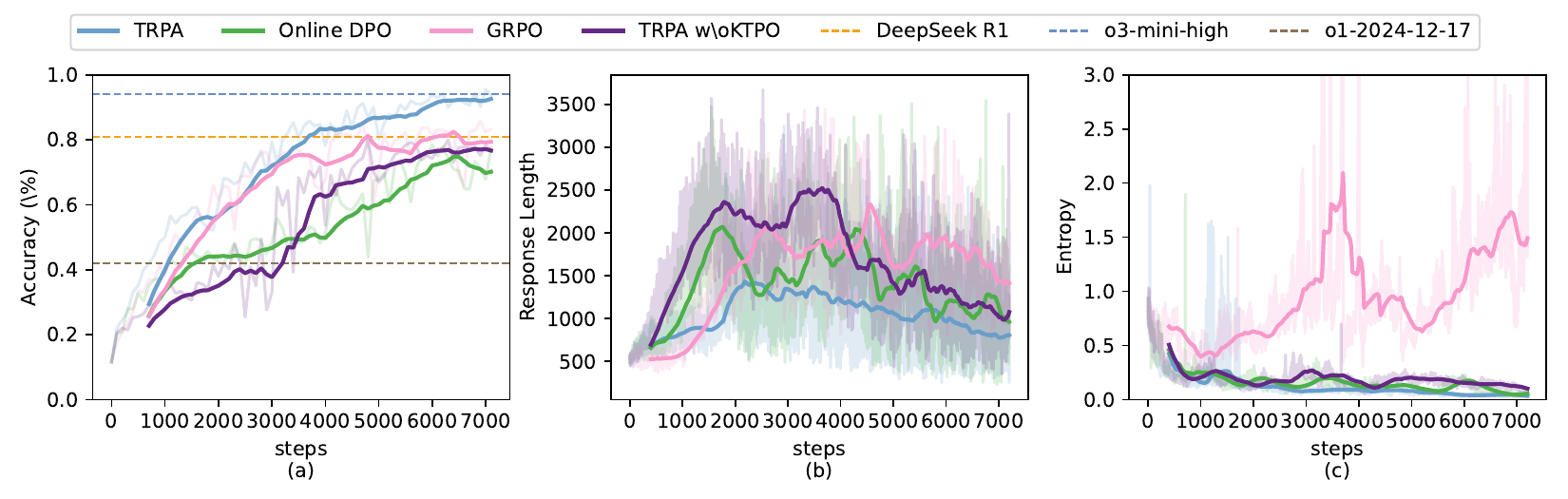}}
\vspace{-0.1in}
\caption{\centering Comparison of TRPA (Blue), TRPA w/oKTPO (Purple), Online DPO (Green) and GRPO (Pink) in terms of (a) Accuracy, (b) Response Length, and (c) Entropy (Response Length and Entropy are averaged by sliding window = 400).}
\label{ab}
\end{center}
\vspace{-0.25in}
\end{figure}

\textbf{RQ1: How does TRPA perform differently compared to GRPO?} In Figure \ref{ab}, the blue and pink solid lines denote the performance metrics of the TRPA and GRPO algorithms, respectively. As shown in Figure \ref{ab}(a), TRPA eventually converges to a performance level comparable to \texttt{o3-mini-high} (blue dashed line), while GRPO performs worse, reaching only the level of \texttt{DeepSeek R1} (orange dashed line). For further analysis of response length and entropy, please refer to \textbf{RQ3}. 

\textbf{RQ2: What benefits does the KL divergence used by TRPA bring compared to Online DPO?} Compared to Online DPO, TRPA demonstrates higher efficiency. In Figure \ref{ab}, the green solid line represents the results of Online DPO. Throughout training, TRPA consistently maintains a lower response length. This advantage allows TRPA to achieve the same training objectives with fewer tokens and reduced computational overhead. Thus, with monotonic improvement guarantee, TRPA exhibits greater optimization efficiency than Online DPO. Additionally, the lower response length contributes to higher token efficiency in the final deployed TRPA-based LLMs.

\textbf{RQ3: The trends of Response Length and Entropy as RL training progresses.} 
Consistent with \cite{deepscaler2025}, the response length of GRPO increases first and then gradually decreases. For the preference-based optimization algorithm (TRPA and Online DPO), we observe a more obvious response length decrease in the later stages of training (Figure \ref{ab}.(b)). It first performs exploration, increasing the response length, and then executes exploitation to adjust response length into a reasonable range. This suggests that, compared to reward-based optimization method GRPO, TRPA exhibits greater advantages in managing response length without requiring additional technical mechanisms, like length penalty. 
Regarding entropy, unlike GRPO, TRPA and Online DPO exhibit greater stability (Figure \ref{ab}.(c)). Stable entropy is generally considered a positive indicator for RL training \cite{li2020suphx}. In the later stages of training, these two preference-based algorithms maintain entropy fluctuations within the range of 0.1 to 0.01, whereas GRPO exhibits more pronounced oscillations.


\textbf{RQ4: The role of KTPO technique.} To evaluate the effectiveness of KTPO, we conduct an ablation study by setting the Kahneman-Tversky factor $N$ to 1 so that the hyperparameter $\beta$ remains uniform across all preference levels. This variant, denoted as TRPA w/o KTPO (purple line in Figure \ref{ab}), reveals several notable differences. In Figure \ref{ab}(a), TRPA w/o KTPO converges to a lower final accuracy than GRPO, indicating diminished optimization effectiveness. Figure \ref{ab}(b) shows that during the mid-training phase (1.5k–3.5k steps), its response length surpasses that of GRPO, while the corresponding accuracy increases only marginally—suggesting that the additional exploration fails to translate into performance gains. After 4.5k steps, its response length declines to a level between that of TRPA and GRPO, indicating a partial retention of TRPA’s length control capability. As shown in Figure \ref{ab}(c), the entropy of TRPA w/o KTPO stabilizes in the range $[0, 0.3]$ after 1k steps—higher than GRPO, indicating improved stability, but still less stable than the full TRPA.

\begin{figure}[ht]
\vskip -0.1in
\begin{center}
\centerline{\includegraphics[width=0.95\columnwidth]{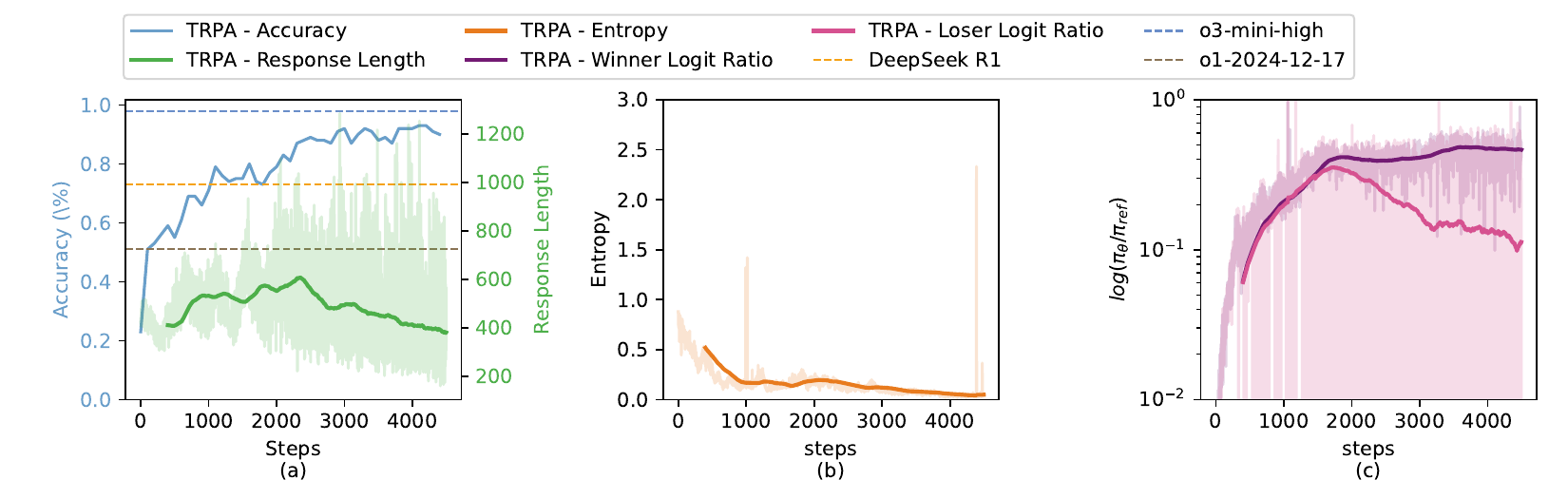}}
\vspace{-0.15in}
\caption{\centering The experiment on K\&K logic puzzle dataset with number of people = 3. (a). Accuracy and Response Length, (b) Entropy, and (c). Logit Ratio $\log(\pi_\theta/\pi_{\text{ref}})$.}
\label{ab2}
\end{center}
\vskip -0.2in
\end{figure}

\textbf{RQ5: Does TRPA have the simultaneous change problem that preference optimization usually faces?}  In Human Alignment tasks, preference-based optimization methods often suffer from the ``simultaneously increase or decrease'' problem, where the absolute probabilities of both preferred and dispreferred responses increase or decrease together \cite{xie2024minor}. This occurs because minimizing the preference loss encourages a higher relative likelihood for preferred responses, which does not necessarily ensure desirable changes in their absolute probabilities \cite{zhang2024automated,pal2024smaug}. 

In contrast, our experiments show that TRPA naturally mitigates this phenomenon. On the K\&K logic puzzle dataset (with three participants), we analyze the logit ratios of preferred (winner) and dispreferred (loser) responses, defined as $\log(\pi_\theta / \pi_{\text{ref}})$, as shown in Figure \ref{ab2}(c). Prior to 1.6k training steps, the logit ratios of both winners and losers increase simultaneously. However, after 1.6k steps, the logit ratio of loser responses begins to decline while that of winners continues to rise, indicating a departure from the simultaneous change pattern. 

We attribute this behavior to TRPA’s online data sampling mechanism. Theoretically, when samples are drawn from the target preference distribution $p^*$, the expected logit ratio of losers decreases while that of winners increases. As training progresses, TRPA's policy $\pi_\theta$ converges to the target distribution $\bar{\pi}^\tau$, then the resulting preference distribution $p_\theta$ converges to $p^*$, thereby avoiding simultaneous changes. 
In contrast, static datasets used in conventional Human Alignment methods lack this adaptivity, making them more susceptible to such problems.

\section{Conclusion, Discussion and Future Work}
In this paper, we introduce the Trust Region Preference Approximation (TRPA) algorithm, an RL framework that integrates rule-based and preference-based optimization to enhance the reasoning capabilities of LLMs. By eliminating the need for reward function, TRPA mitigates reward hacking while ensuring Monotonic Improvement Guarantees toward the target distribution $\bar{\pi}^\tau$. Experimental results validate its effectiveness, demonstrating superior reasoning performance and training stability. These findings suggest that TRPA offers a promising direction for improving large-scale RL algorithms in LLM reasoning. 

The simplicity and stability of TRPA make it highly promising for applications in scientific research. This paper serves as an early exploration of foundational algorithms in the AI for Science domain. We are actively working on extending this approach to key scientific research areas, such as drug discovery and molecular retrosynthesis. Our future work will conduct a more detailed study of algorithms such as TRPA based on some classic problems that are usually concerned in the RL field:

\ding{192} \textbf{UTD (Update-to-Data) ratio.} Even if we have already deployed multiple optimization steps on the off-policy data $\mathcal{D}_{\text{old}}$, we still observe that some preference pairs with good enough quality were not fully optimized but were discarded. Therefore, introducing technologies such as the Replay Buffer \cite{mnih2015human}, Prioritized Experience Replay (PER) \cite{schaul2015prioritized} and Model-Based Methods \cite{janner2019trust} to enhance data utilization is helpful for enhancing the ability of LLM reasoning and planning. In other words, the current optimization of LLMs' policy on existing data is still insufficient, which may also be the reason why the current use of RL algorithms does not trigger the Plasticity Loss phenomenon \cite{abbas2023loss}, which is often encountered in Deep RL and Continual Learning.

\ding{193} \textbf{Exploration-Exploitation.} About RL exploration, the increase in response length can be understood as the exploration of the policy in RL. For wrong questions, more exploration should be conducted so that LLM can use more tokens for exploration. For correct questions, exploration should focus on how to solve the problem more efficiently. Regarding exploitation, the utilization patterns of low-variance data and high-variance data should be different. Low-variance data means that LLM tends to get it right most of the time or get it wrong most of the time. Such data is less useful for exploitation of RL training. Our TRPA processing at the response level is to more directly use the comparison of logits between different responses. The logit of the correct answer should become larger, otherwise it will become smaller. In this process, the logits of different answers will participate in each other's gradient calculations, while in the GRPO algorithm, only different advantages (not logits) will participate in the gradient calculations.

\ding{194} \textbf{Learning Signal Sparsity.} For current large-scale RL algorithms, the learning signal comes from the evaluation of responses based on reward function or preference levels defined by rules. Therefore, the sequence-level learning signal is still sparse for token-level optimization. How to design richer token-level or reasoning-step-level learning signals is the key idea to enhance algorithm performance.

\ding{195} \textbf{RL Cold Start.} The performance of the base model determines the upper limit of post-training technology such as large-scale RL. However, it is still possible to further break through the original performance upper limit through the design of post-training technology, such as designing a multi-stage RL algorithm to alleviate the phenomenon of language mixing or enhancing the necessary knowledge mastery and logical thinking ability when LLMs reasoning.



\bibliographystyle{unsrt}  
\bibliography{references}

\newpage
\appendix



\section{Discussion of Preference-based Optimization, Reward-based Optimization and Rule-based Optimization methods}\label{sec:method_discussion}
The basic process of the three methods is shown in Figure \ref{ROROPO_TRPA}. Here we focus on the relationship and difference among these methods. 
These methods have overlaps with each other but are different from each other. Firstly, the recently proposed DeepSeek R1 \cite{DeepSeekR1} proposed that using Reward-based Optimization method (such as GRPO) for Rule-based optimization, where the reward function comes from predefined rules (such as whether the correct answer is output). But not all reward functions can be abstracted from a set of predefined rules. For example, \cite{jaderberg2018human} proposed that in complex games, the reward functions to optimize the intelligent agent to reach the human level should be automatically evolutionary. Secondly, in addition to Preference-based and Reward-based Optimization, Rule-based Optimization also includes other types of methods, such as model-based optimization, evolution-based optimization (e.g. Ant Colony Optimization (ACO) \cite{dorigo1996ant}) methods, etc. These methods optimize model behavior through other mechanisms (implicit rules). For example, model-based policy optimization methods \cite{janner2019trust,zhu2021making,moerland2023model} optimize by building an environmental model to predict future rewards. Thirdly, Preference-based Optimization such as human preference alignment cannot usually be modeled as Rule-based Optimization. This is because human preferences are unclear, changing, and have individual differences, making it difficult to abstract them into unified rules. Only some specific tasks such as safety can be designed as rule-based optimization-like methods \cite{mu2025rule}, but they still need to use artificial intelligence models to model complex safety guidelines. Our paper intends to design a new type of Preference-based Optimization method based on the idea of Rule-based Optimization to enhance the reasoning capability of LLMs.

Here we introduce several RL algorithms for Reward-based Optimization that are commonly used in LLMs Reasoning tasks.

\paragraph{Proximal Policy Optimization (PPO).} In the response generation process, a language model is prompted with an input question $x$ to generate an output $y$, where both $x$ and $y$ are sequences of tokens. Proximal Policy Optimization (PPO) \cite{PPO-basedRLHF} algorithm incorporates a Kullback-Leibler (KL) divergence penalty to constrain the updated policy $\pi_\theta$ from deviating too far from a reference policy $\pi_{\text{ref}}$. The optimization objective for the policy $\pi_\theta$ can be formulated as:
\begin{equation}
    \begin{aligned}
        \label{eq:kl}
        \mathbb{E}_{x \sim D, y \sim \pi_\theta(\cdot \mid x)}\left[r_\phi(x, y)\right] - \beta D_{\mathrm{KL}}\left(\pi_\theta(\cdot \mid x) \| \pi_{\mathrm{ref}}(\cdot \mid x)\right)\,,
    \end{aligned}
\end{equation}
where $\beta > 0$ is a hyperparameter controlling the regularization strength.

\paragraph{Group Relative Policy Optimization (GRPO).} GRPO \cite{shao2024deepseekmath} is a reinforcement learning algorithm designed to optimize policies by leveraging group-wise relative preference information. Instead of relying on absolute reward values, GRPO updates policies based on relative comparisons within predefined groups, improving stability and sample efficiency.
\begin{equation}
\small{\begin{aligned}
\mathcal{L}_{\text{GRPO}}(\theta)=&\mathbb{E}\left[x\sim \mathcal{D},\left\{y_i\right\}_{i=1}^G \sim \pi_{\theta_{\text{old}}}(\cdot \mid x)\right] \\
\frac{1}{G} \sum_{i=1}^G&\left(\min \left(\frac{\pi_\theta\left(y_i \mid x\right)}{\pi_{\theta_{\text{old}}}\left(y_i \mid x\right)} A_i, \operatorname{clip}\left(\frac{\pi_\theta\left(y_i \mid x\right)}{\pi_{\theta_{\text{old}}}\left(y_i \mid x\right)}, 1-\varepsilon, 1+\varepsilon\right) A_i\right)-\beta \mathbb{D}_{K L}\left(\pi_\theta| | \pi_{r e f}\right)\right),
\end{aligned}}
\end{equation}
where the advantage $A_i=\frac{r_i-\operatorname{mean}\left(\left\{r_1, r_2, \cdots, r_G\right\}\right)}{\operatorname{std}\left(\left\{r_1, r_2, \cdots, r_G\right\}\right)}$ is computed using a group of rewards.

\section{Theorem Proof}
\subsection{Proof of Lemma \ref{Online_DPO_is_not_PBA}}\label{Online_DPO_is_not_PBA_proof}
Lemma \ref{Online_DPO_is_not_PBA}: The Online DPO algorithm is not Posterior Boltzmann Approximation (PBA) algorithm.

Proof: Recall that
\begin{equation}
    \begin{aligned}
&\mathcal{L}_{\mathrm{Online\ DPO}}\left(\pi_{\theta} ; \pi_{\text {ref }}\right)=-\mathbb{E}_{\left(x, y_1, y_2\right) \sim \mathcal{D}_\theta}\left[\log \sigma\left(\bar{h}_{\theta}\left(x, y_1, y_2\right)\right)\right]\\
=&-\mathbb{E}_{x\sim \mathcal{D},y_1,y_2\sim \pi_\theta(y|x)}[p^*(z=1|y_1,y_2,x)\log \sigma\left(\bar{h}_\theta(x,y_1,y_2)\right) \\
&\quad+ p^*(z=0|y_1,y_2,x)\log \sigma\left(\bar{h}_\theta (x,y_2,y_1)\right)] \\
=&\mathbb{E}_{x\sim \mathcal{D},y_1,y_2\sim \pi_\theta(y|x)}\left[\mathrm{H}(p^*(z|y_1,y_2,x)||p_\theta(z|y_1,y_2,x))\right].
\end{aligned}
\end{equation}

Calculate the gradient of $\mathcal{L}_{\mathrm{Online\ DPO}}\left(\pi_{\theta} ; \pi_{\text {ref }}\right)$ about $\theta$:
\begin{equation}
\begin{aligned}
    &\nabla_\theta\mathcal{L}_{\mathrm{Online\ DPO}}\left(\pi_{\theta} ; \pi_{\text {ref }}\right)=\nabla_\theta\mathbb{E}_{x\sim \mathcal{D},y_1,y_2\sim \pi_\theta(y|x)}\left[\mathrm{H}(p^*(z|y_1,y_2,x)||p_\theta(z|y_1,y_2,x))\right]\\
    =&\int \mathcal{D}(x) (\mathrm{H}(p^*(z|y_1,y_2,x)||p_\theta(z|y_1,y_2,x))\nabla_\theta (\pi_\theta(y_1|x)\pi_\theta(y_2|x))\\
    &\quad\quad\quad+ \pi_\theta(y_1|x)\pi_\theta(y_2|x)\nabla_\theta \mathrm{H}(p^*(z|y_1,y_2,x)||p_\theta(z|y_1,y_2,x)))dy_1dy_2dx.
\end{aligned}
\end{equation}
As for $\nabla_\theta \mathrm{H}(p^*(z|y_1,y_2,x)||p_\theta(z|y_1,y_2,x))$, we have 
\begin{equation}
\begin{aligned}\label{nabla_H}
    \nabla_\theta \mathrm{H}(p^*(z|y_1,y_2,x)||p_\theta(z|y_1,y_2,x)) =& -\left(\frac{p^*(z|y_1,y_2,x)}{p_\theta(z|y_1,y_2,x)}\right. \\
    &\quad\quad- \left.\frac{1-p^*(z|y_1,y_2,x)}{1-p_\theta(z|y_1,y_2,x)}\right)\nabla_\theta p_\theta(z|y_1,y_2,x) .
\end{aligned}
\end{equation}
Consider using the BT model \cite{bradley1952rank}, we have that when $\pi_\theta$ is $\bar{\pi}^\tau(y|x)$, because $p_\theta(z=1|y_1,y_2,x)=\sigma\left(\bar{h}_\theta(x,y_1,y_2)\right)$ and $p_\theta(z=0|y_1,y_2,x)=\sigma\left(\bar{h}_\theta(x,y_2,y_1)\right)$, then
\begin{equation}\label{pstar_equal_ptheta}
\begin{aligned}
&p_\theta(z=1|y_1,y_2,x)=\sigma\left(\bar{h}_\theta(x,y_1,y_2)\right) \\
=&\sigma\left(\beta \log \frac{\pi_\theta\left(y_1 \mid x\right)}{\pi_{\text {ref }}\left(y_1 \mid x\right)}-\beta \log \frac{\pi_\theta\left(y_2 \mid x\right)}{\pi_{\text {ref }}\left(y_2 \mid x\right)}\right)=\sigma\left(r(x,y_1)-r(x,y_2)\right) \\
=&\frac{\exp \left(r\left(x, y_1\right)\right)}{\exp \left(r\left(x, y_1\right)\right)+\exp \left(r\left(x, y_2\right)\right)} = p^*(z=1|y_1,y_2,x).
\end{aligned}
\end{equation}
And similarly, $p_\theta(z=0|y_1,y_2,x)=p^*(z=0|y_1,y_2,x)$. Then for Eq.\ref{nabla_H}, we have
\begin{equation}
    \nabla_\theta \mathrm{H}(p^*(z|y_1,y_2,x)||p_\theta(z|y_1,y_2,x)) =0.
\end{equation}

But for $\mathrm{H}(p^*(z|y_1,y_2,x)||p_\theta(z|y_1,y_2,x))$, as shown in Figure \ref{hpq_kl_figure} (a), (c), it will not be 0 except $p^*(z|y_1,y_2,x)$, $p_\theta(z|y_1,y_2,x)$ are equal to 0 or 1 which usually does not appear in comparison models such as BT model. Then for $\nabla_\theta\mathcal{L}_{\mathrm{Online\ DPO}}\left(\pi_{\theta} ; \pi_{\text {ref }}\right)$, it would not be zero all the time when $\pi_\theta$ is $\bar{\pi}^\tau(y|x)$, i.e. $\bar{\pi}^\tau(y|x)$ is not the target distribution. Therefore, the Online DPO algorithm is not Posterior Boltzmann Approximation (PBA) algorithm. Proof finished. $\hfill\qedsymbol$

\subsection{Proof of Lemma \ref{TRPA_is_PBA}}\label{TRPA_is_PBA_proof}
Lemma \ref{TRPA_is_PBA}: The PA algorithm is Posterior Boltzmann Approximation (PBA) algorithm.

Proof: Recall
\begin{equation}
\begin{aligned}
    \mathcal{L}_{\mathrm{PA}}(\pi_\theta; \pi_{\text {ref }})
&=-\mathbb{E}_{(x,y_1,y_2)\sim \mathcal{D}_{\theta}}\left[\log \sigma\left(\bar{h}_\theta\left(x, y_1, y_2\right)\right)\right]+\mathbb{E}_{x\sim \mathcal{D},y_1,y_2\sim \pi_\theta(y|x)}\left[M(x,y_1,y_2)\right] \\
&=\mathbb{E}_{x\sim \mathcal{D},y_1,y_2\sim \pi_\theta(y|x)}\left[\mathrm{D}_{\mathrm{KL}}(p^*(z|y_1,y_2,x)||p_\theta(z|y_1,y_2,x))\right].
\end{aligned}
\end{equation}
Calculate the gradient of $\mathcal{L}_{\mathrm{PA}}(\pi_\theta; \pi_{\text {ref }})$ about $\theta$:
\begin{equation}
\begin{aligned}
    &\nabla_\theta\mathcal{L}_{\mathrm{PA}}(\pi_\theta; \pi_{\text {ref }})=\nabla_\theta\mathbb{E}_{x\sim \mathcal{D},y_1,y_2\sim \pi_\theta(y|x)}\left[\mathrm{D}_{\mathrm{KL}}(p^*(z|y_1,y_2,x)||p_\theta(z|y_1,y_2,x))\right]\\
    =&\int \mathcal{D}(x) (\mathrm{D}_{\mathrm{KL}}(p^*(z|y_1,y_2,x)||p_\theta(z|y_1,y_2,x))\nabla_\theta (\pi_\theta(y_1|x)\pi_\theta(y_2|x))\\
    &\quad\quad\quad+ \pi_\theta(y_1|x)\pi_\theta(y_2|x)\nabla_\theta \mathrm{D}_{\mathrm{KL}}(p^*(z|y_1,y_2,x)||p_\theta(z|y_1,y_2,x)))dy_1dy_2dx.
\end{aligned}
\end{equation}
Because $\mathrm{D}_{\mathrm{KL}}(p||q)=\mathrm{E}_p[\log p-\log q]$, we have that:
\begin{equation}
\nabla_q \mathrm{D}_{\mathrm{KL}}(p||q)=\nabla_q\mathrm{E}_p[\log p-\log q] = -\nabla_q \mathrm{E}_p[\log q] = \nabla_q \mathrm{H}(p||q).
\end{equation}
Consider using the BT model \cite{bradley1952rank}, we have that when $\pi_\theta$ is $\bar{\pi}^\tau(y|x)$, based on Eq.\ref{nabla_H}, then
\begin{equation}
\begin{aligned}
    \nabla_\theta \mathrm{D}_{\mathrm{KL}}(p^*(z|y_1,y_2,x)||p_\theta(z|y_1,y_2,x)) =& -\left(\frac{p^*(z|y_1,y_2,x)}{p_\theta(z|y_1,y_2,x)} \right. \\
    &\quad\quad- \left. \frac{1-p^*(z|y_1,y_2,x)}{1-p_\theta(z|y_1,y_2,x)}\right)\nabla_\theta p_\theta(z|y_1,y_2,x) .
\end{aligned}
\end{equation}

Based on Eq.\ref{pstar_equal_ptheta}, we have $p_\theta(z|y_1,y_2,x)=p^*(z|y_1,y_2,x)$. Then 
\begin{equation}
    \nabla_\theta \mathrm{D}_{\mathrm{KL}}(p^*(z|y_1,y_2,x)||p_\theta(z|y_1,y_2,x))=0.
\end{equation}

And as a well-known conclusion, $\mathrm{D}_{\mathrm{KL}}(p^*(z|y_1,y_2,x)||p_\theta(z|y_1,y_2,x))=0$ when $p_\theta(z|y_1,y_2,x)=p^*(z|y_1,y_2,x)$. Then we get that $\nabla_\theta\mathcal{L}_{\mathrm{PA}}(\pi_\theta; \pi_{\text {ref }})=0$ when $\pi_\theta$ is $\bar{\pi}^\tau(y|x)$. Therefore, the PA algorithm is Posterior Boltzmann Approximation (PBA) algorithm. Proof finished. $\hfill\qedsymbol$

\subsection{Proof of Theorem \ref{TRPA} }\label{TRPRA_bound_proof}
\textbf{Theorem \ref{TRPA}:} Denote 
    $$\begin{aligned}
        L^{\pi_{\theta_{\text{old}}}}(\pi_{\theta_{\text{new}}})=&-\mathbb{E}_{(x,y_1,y_2)\sim \mathcal{D}_{{\text{old}}}}\left[\log\sigma\left( \bar{h}_{\theta_{\text{new}}}\left(x, y_1, y_2\right)\right)\right]+\mathbb{E}_{x\sim \mathcal{D},y_1,y_2\sim \pi_{\theta_{\text{old}}}(y|x)}\left[M(x,y_1,y_2)\right].
    \end{aligned}$$
Then we have the following inequality hold:
\begin{equation}
    \small{
    \begin{aligned}
        L^{\pi_{\theta_{\text{new}}}}(\pi_{\theta_{\text{new}}}) \leq &L^{\pi_{\theta_{\text{old}}}}(\pi_{\theta_{\text{new}}}) + a_1\sqrt{D_{\mathrm{KL}}^{\max }(\pi_{\theta_{\text{old}}}|| \pi_{\theta_{\text{new}}})}.
    \end{aligned}
    }
\end{equation}
where $a_1=4(\bar{U}_r + 2\log2)$, $\bar{U}_r=\max_{x,y_1,y_2}  \mathrm{D}_{\mathrm{KL}}(p^*(z|y_1,y_2,x)||p_{\theta_{\text{new}}}(z|x,y_1,y_2))$ and $M(x,y_1,y_2)=\sum_{z=0,1}p^*(z|y_1,y_2,x)\log p^*(z|y_1,y_2,x)$.

\textbf{Proof:} First we propose some Lemma:
\begin{lemma}\label{SelfEntropy}
    For $f(p)=-p\log p-(1-p)\log(1-p)$, $f(p)\leq\log2$.

    Proof: For $f(p)=-p\log p-(1-p)\log(1-p)\geq 0, p\in[0,1]$, $f'(p)=\log(\frac{1}{p}-1)$ is a monotonically decreasing function that reaches 0 when $p=0.5$. Thus $f(p)$ increases first and then decreases, and reaches its maximum value at $p=0.5$. So $f(p)\leq\log2$.
\end{lemma}

\begin{lemma}\label{TV_KL_inequal}
$D_{\mathrm{TV}}(p \| q)^2 \leq D_{\mathrm{KL}}(p \| q)$.
\end{lemma}

Consider the upper bound of $|L^{\pi_{\theta_{\text{new}}}}(\pi_{\theta_{\text{new}}})-L^{\pi_{\theta_{\text{old}}}}(\pi_{\theta_{\text{new}}})|$:
\begin{equation}\label{PRA_bound_eq1}\small{
    \begin{aligned}
&|L^{\pi_{\theta_{\text{new}}}}(\pi_{\theta_{\text{new}}})-L^{\pi_{\theta_{\text{old}}}}(\pi_{\theta_{\text{new}}})|\\
=&|-\mathbb{E}_{x\sim \mathcal{D}}\sum_{y_1,y_2\in\mathcal{Y}}[\pi_{\theta_{\text{new}}}(y_1|x)\pi_{\theta_{\text{new}}}(y_2|x)(p^*(1|y_1,y_2,x)\log\sigma\left( h_{\theta_{\text{new}}}(x,y_1,y_2)\right)\\
&\quad +p^*(0|y_1,y_2,x)\log\sigma\left( h_{\theta_{\text{new}}}(x,y_2,y_1)\right))]\\
&+\mathbb{E}_{x\sim \mathcal{D}}\sum_{y_1,y_2\in\mathcal{Y}}\left[(\pi_{\theta_{\text{new}}}(y_1|x)\pi_{\theta_{\text{new}}}(y_2|x)-\pi_{\theta_{\text{old}}}(y_1|x)\pi_{\theta_{\text{old}}}(y_2|x))M(x,y_1,y_2)\right]|\\
\leq&|-\mathbb{E}_{x\sim \mathcal{D}}\sum_{y_1,y_2\in\mathcal{Y}}[\pi_{\theta_{\text{new}}}(y_1|x)\pi_{\theta_{\text{new}}}(y_2|x)(p^*(1|y_1,y_2,x)\log\sigma\left( h_{\theta_{\text{new}}}(x,y_1,y_2)\right)\\
&\quad +p^*(0|y_1,y_2,x)\log\sigma\left( h_{\theta_{\text{new}}}(x,y_2,y_1)\right))]|\\
&+|\mathbb{E}_{x\sim \mathcal{D}}\sum_{y_1,y_2\in\mathcal{Y}}\left[(\pi_{\theta_{\text{new}}}(y_1|x)\pi_{\theta_{\text{new}}}(y_2|x)-\pi_{\theta_{\text{old}}}(y_1|x)\pi_{\theta_{\text{old}}}(y_2|x))M(x,y_1,y_2)\right]|.
    \end{aligned}
}
\end{equation}

Define $D_{T V}(p \| q)=\frac{1}{2} \sum_i\left|p_i-q_i\right|$ for discrete probability distributions $p, q$ and define $D_{\mathrm{TV}}^{\max }(\pi_{\theta_{\text{new}}}|| \pi_{\theta_{\text{old}}})$ as:
\begin{equation}
    D_{\mathrm{TV}}^{\max }(\pi_{\theta_{\text{new}}}|| \pi_{\theta_{\text{old}}})=\max _{x\in\mathcal{X}} D_{T V}(\pi_{\theta_{\text{new}}}(\cdot \mid x) \| \pi_{\theta_{\text{old}}}(\cdot \mid x)).
\end{equation}
For the term $|\mathbb{E}_{x\sim \mathcal{D}}\sum_{y_1,y_2\in\mathcal{Y}}\left[(\pi_{\theta_{\text{new}}}(y_1|x)\pi_{\theta_{\text{new}}}(y_2|x)-\pi_{\theta_{\text{old}}}(y_1|x)\pi_{\theta_{\text{old}}}(y_2|x))M(x,y_1,y_2)\right]|$, we have:
\begin{equation}\label{TV_distance_proof}\small{
    \begin{aligned}
        &|\mathbb{E}_{x\sim \mathcal{D}}\sum_{y_1,y_2\in\mathcal{Y}}\left[(\pi_{\theta_{\text{new}}}(y_1|x)\pi_{\theta_{\text{new}}}(y_2|x)-\pi_{\theta_{\text{old}}}(y_1|x)\pi_{\theta_{\text{old}}}(y_2|x))M(x,y_1,y_2)\right]|\\
        \leq&\mathbb{E}_{x\sim \mathcal{D}}\sum_{y_1,y_2\in\mathcal{Y}}\left[|(\pi_{\theta_{\text{new}}}(y_1|x)\pi_{\theta_{\text{new}}}(y_2|x)-\pi_{\theta_{\text{old}}}(y_1|x)\pi_{\theta_{\text{old}}}(y_2|x))|\cdot|M(x,y_1,y_2)|\right]\\
        \leq& \mathbb{E}_{x\sim \mathcal{D}}\sum_{y_1,y_2\in\mathcal{Y}}\left[|(\pi_{\theta_{\text{new}}}(y_1|x)\pi_{\theta_{\text{new}}}(y_2|x)-\pi_{\theta_{\text{old}}}(y_1|x)\pi_{\theta_{\text{old}}}(y_2|x))|\log2\right]\\
        \leq& \mathbb{E}_{x\sim \mathcal{D}}\sum_{y_1,y_2\in\mathcal{Y}}[|\pi_{\theta_{\text{new}}}(y_1|x)\pi_{\theta_{\text{new}}}(y_2|x)-\pi_{\theta_{\text{new}}}(y_1|x)\pi_{\theta_{\text{old}}}(y_2|x)| \\
        &\quad\quad\quad\quad\quad\quad+|\pi_{\theta_{\text{new}}}(y_1|x)\pi_{\theta_{\text{old}}}(y_2|x) -\pi_{\theta_{\text{old}}}(y_1|x)\pi_{\theta_{\text{old}}}(y_2|x)|]\log2\\
    \end{aligned}
}
\end{equation}
\begin{equation}\small{
    \begin{aligned}
        =&\mathbb{E}_{x\sim \mathcal{D}}\sum_{y_1,y_2\in\mathcal{Y}}[|\pi_{\theta_{\text{new}}}(y_1|x)(\pi_{\theta_{\text{new}}}(y_2|x)-\pi_{\theta_{\text{old}}}(y_2|x))|\\
        &\quad\quad\quad\quad\quad\quad+|(\pi_{\theta_{\text{new}}}(y_1|x) -\pi_{\theta_{\text{old}}}(y_1|x))\pi_{\theta_{\text{old}}}(y_2|x)|]\log2\\
        =&\mathbb{E}_{x\sim \mathcal{D}}\sum_{y_1,y_2\in\mathcal{Y}}[\pi_{\theta_{\text{new}}}(y_1|x)\cdot|(\pi_{\theta_{\text{new}}}(y_2|x)-\pi_{\theta_{\text{old}}}(y_2|x))| \\
        &\quad\quad\quad\quad\quad\quad+|(\pi_{\theta_{\text{new}}}(y_1|x) -\pi_{\theta_{\text{old}}}(y_1|x))|\cdot\pi_{\theta_{\text{old}}}(y_2|x)]\log2\\
        =& \mathbb{E}_{x\sim \mathcal{D}}[\sum_{y_2\in\mathcal{Y}}(\sum_{y_1\in\mathcal{Y}}\pi_{\theta_{\text{new}}}(y_1|x))\cdot|(\pi_{\theta_{\text{new}}}(y_2|x)-\pi_{\theta_{\text{old}}}(y_2|x))| \\
        &\quad\quad\quad\quad\quad\quad+\sum_{y_1\in\mathcal{Y}}|(\pi_{\theta_{\text{new}}}(y_1|x) -\pi_{\theta_{\text{old}}}(y_1|x))|\cdot(\sum_{y_2\in\mathcal{Y}}\pi_{\theta_{\text{old}}}(y_2|x))]\log2\\
        =& \mathbb{E}_{x\sim \mathcal{D}}[\sum_{y_2\in\mathcal{Y}}|(\pi_{\theta_{\text{new}}}(y_2|x)-\pi_{\theta_{\text{old}}}(y_2|x))|+\sum_{y_1\in\mathcal{Y}}|(\pi_{\theta_{\text{new}}}(y_1|x) -\pi_{\theta_{\text{old}}}(y_1|x))|]\log2\\
        = &\mathbb{E}_{x\sim \mathcal{D}}[2D_{T V}(\pi_{\theta_{\text{new}}}(\cdot \mid x) \| \pi_{\theta_{\text{old}}}(\cdot \mid x))+2D_{T V}(\pi_{\theta_{\text{new}}}(\cdot \mid x) \| \pi_{\theta_{\text{old}}}(\cdot \mid x))]\log2\\
        =&4\log2\cdot D_{\mathrm{TV}}^{\max }(\pi_{\theta_{\text{new}}}|| \pi_{\theta_{\text{old}}}).
    \end{aligned}
}
\end{equation}
The second inequality holds as $|M(x,y_1,y_2)|=|\sum_{z=0,1}p^*(z|y_1,y_2,x)\log p^*(z|y_1,y_2,x)|\leq \log2$ cause Lemma \ref{SelfEntropy}.
\begin{equation}
    \begin{aligned}
        &|M(x,y_1,y_2)|=|\sum_{z=0,1}p^*(z|y_1,y_2,x)\log p^*(z|y_1,y_2,x)|\\
        =&|p^*(z=1|y_1,y_2,x)\log p^*(z=1|y_1,y_2,x)+p^*(z=0|y_1,y_2,x)\log p^*(z=0|y_1,y_2,x)|\\
        =&|-p^*(z=1|y_1,y_2,x)\log p^*(z=1|y_1,y_2,x) \\
        &\quad- (1-p^*(z=1|y_1,y_2,x))\log (1-p^*(z=1|y_1,y_2,x))|.
    \end{aligned}
\end{equation}

Denote $M_1(x,y_1,y_2)=(p^*(z=1|y_1,y_2,x)\log\sigma\left( \bar{h}_{\theta_{\text{new}}}(x,y_1,y_2)\right)+p^*(z=0|y_1,y_2,x)\log\sigma\left( \bar{h}_{\theta_{\text{new}}}(x,y_2,y_1)\right))$. Then for the first term in Eq.\ref{PRA_bound_eq1}, we have:
\begin{equation}\label{PRA_bound_eq2}
\small{
    \begin{aligned}
        &|-\mathbb{E}_{x\sim \mathcal{D}}\sum_{y_1,y_2\in\mathcal{Y}}[\pi_{\theta_{\text{new}}}(y_1|x)\pi_{\theta_{\text{new}}}(y_2|x)(p^*(1|y_1,y_2,x)\log\sigma\left( \bar{h}_{\theta_{\text{new}}}(x,y_1,y_2)\right) \\
        &+p^*(0|y_1,y_2,x)\log\sigma\left( \bar{h}_{\theta_{\text{new}}}(x,y_2,y_1)\right))]|\\
=&|\mathbb{E}_{x\sim \mathcal{D}}\sum_{y_1,y_2\in\mathcal{Y}}[-\pi_{\theta_{\text{new}}}(y_1|x)\pi_{\theta_{\text{new}}}(y_2|x)M_1(x,y_1,y_2)+ \pi_{\theta_{\text{old}}}(y_1|x)\pi_{\theta_{\text{old}}}(y_2|x)M_1(x,y_1,y_2)]|\\
\leq& \mathbb{E}_{x\sim \mathcal{D}}\sum_{y_1,y_2\in\mathcal{Y}}\left[|\pi_{\theta_{\text{new}}}(y_1|x)(\pi_{\theta_{\text{new}}}(y_2|x)-\pi_{\theta_{\text{old}}}(y_2|x))|\right. \\
&\quad\quad\quad\quad\quad\quad\left.+|(\pi_{\theta_{\text{new}}}(y_1|x) -\pi_{\theta_{\text{old}}}(y_1|x))\pi_{\theta_{\text{old}}}(y_2|x)|\right]\cdot\max_{x,y_1,y_2}|M_2(x,y_1,y_2)|
    \end{aligned}
    }
\end{equation}
\begin{equation} 
\small{
    \begin{aligned}
=& \mathbb{E}_{x\sim \mathcal{D}}\left[\sum_{y_2\in\mathcal{Y}}|(\pi_{\theta_{\text{new}}}(y_2|x)-\pi_{\theta_{\text{old}}}(y_2|x))|\right. \\
&\quad\quad\quad\left.+\sum_{y_1\in\mathcal{Y}}|(\pi_{\theta_{\text{new}}}(y_1|x) -\pi_{\theta_{\text{old}}}(y_1|x))|\right]\cdot\max_{x,y_1,y_2}|M_2(x,y_1,y_2)|\\
= &\mathbb{E}_{x\sim \mathcal{D}}[2D_{T V}(\pi_{\theta_{\text{new}}}(\cdot \mid x) \| \pi_{\theta_{\text{old}}}(\cdot \mid x))+2D_{T V}(\pi_{\theta_{\text{new}}}(\cdot \mid x) \| \pi_{\theta_{\text{old}}}(\cdot \mid x))]\cdot\max_{x,y_1,y_2}|M_2(x,y_1,y_2)|\\
=&4\cdot\max_{x,y_1,y_2}|M_2(x,y_1,y_2)|\cdot D_{\mathrm{TV}}^{\max }(\pi_{\theta_{\text{new}}}|| \pi_{\theta_{\text{old}}}).
    \end{aligned}
    }
\end{equation}
Denote $P=p^*(z=1|y_1,y_2,x), Q=\sigma\left( \bar{h}_{\theta_{\text{new}}}(x,y_1,y_2)\right)$. Define a parameterized conditional 0-1 distribution $p_{\theta_{\text{new}}}(z|y_1,y_2,x)$ where $z\in\{0,1\}$, $p_{\theta_{\text{new}}}(0|y_1,y_2,x)=\sigma\left( \bar{h}_{\theta_{\text{new}}}(x,y_2,y_1)\right)$ and $p_{\theta_{\text{new}}}(1|y_1,y_2,x)=\sigma\left( \bar{h}_{\theta_{\text{new}}}(x,y_1,y_2)\right)$. Denote $\bar{U}_r=\max_{x,y_1,y_2}  \mathrm{D}_{\mathrm{KL}}(p^*(z|y_1,y_2,x)|| p_{\theta_{\text{new}}}(z|x,y_1,y_2))$. 
Now we take $\max_{x,y_1,y_2}|M_2(x,y_1,y_2)|$ as:
\begin{equation}
    \begin{aligned}
        &\max_{x,y_1,y_2}|M_2(x,y_1,y_2)| = \max_{x,y_1,y_2} |-P\log Q + (1-P)\log(1-Q)| \\
        =& \max_{x,y_1,y_2} |P\log\frac{P}{Q} - (1-P)\log\frac{1-P}{1-Q} - P\log P - (1-P)\log(1-P)|\\
        \leq& \max_{x,y_1,y_2} |P\log\frac{P}{Q} - (1-P)\log\frac{1-P}{1-Q} |+\max_{x,y_1,y_2}| P\log P + (1-P)\log(1-P)|\\
        \leq& \max_{x,y_1,y_2} |P\log\frac{P}{Q} - (1-P)\log\frac{1-P}{1-Q} |+ \log2\\
=& \max_{x,y_1,y_2}  \mathrm{D}_{\mathrm{KL}}(p^*(z|y_1,y_2,x)|| p_{\theta_{\text{new}}}(z|x,y_1,y_2)) + \log2\\
        \leq& \max_{x,y_1,y_2}  \mathrm{D}_{\mathrm{KL}}(p^*(z|y_1,y_2,x)|| p_{\theta_{\text{new}}}(z|x,y_1,y_2))  + \log2 = \bar{U}_r + \log2.\\
    \end{aligned}
\end{equation}

The second inequality holds as $|P\log P + (1-P)\log(1-P)|\leq \log2$ cause Lemma \ref{SelfEntropy}.

Then we have:
\begin{equation}
    \begin{aligned}
        &|L^{\pi_{\theta_{\text{new}}}}(\pi_{\theta_{\text{new}}})-L^{\pi_{\theta_{\text{old}}}}(\pi_{\theta_{\text{new}}})|\\
\leq& 4\log2\cdot D_{\mathrm{TV}}^{\max }(\pi_{\theta_{\text{new}}}|| \pi_{\theta_{\text{old}}}) + 4(\bar{U}_r + \log2)\cdot D_{\mathrm{TV}}^{\max }(\pi_{\theta_{\text{new}}}|| \pi_{\theta_{\text{old}}}) \\
=& 4(\bar{U}_r + 2\log2)\cdot D_{\mathrm{TV}}^{\max }(\pi_{\theta_{\text{old}}}|| \pi_{\theta_{\text{new}}})\\
\leq& 4(\bar{U}_r + 2\log2)\sqrt{D_{\mathrm{KL}}^{\max }(\pi_{\theta_{\text{old}}}|| \pi_{\theta_{\text{new}}})}.
    \end{aligned}
\end{equation}
The second inequality holds cause Lemma \ref{TV_KL_inequal}. Then finally we have:
\begin{equation}
    \begin{aligned}
        &L^{\pi_{\theta_{\text{new}}}}(\pi_{\theta_{\text{new}}})-L^{\pi_{\theta_{\text{old}}}}(\pi_{\theta_{\text{new}}})\leq|L^{\pi_{\theta_{\text{new}}}}(\pi_{\theta_{\text{new}}})-L^{\pi_{\theta_{\text{old}}}}(\pi_{\theta_{\text{new}}})|
\leq a_1\sqrt{D_{\mathrm{KL}}^{\max }(\pi_{\theta_{\text{old}}}|| \pi_{\theta_{\text{new}}})}.
    \end{aligned}
\end{equation}
where $a_1=4(\bar{U}_r + 2\log2)$. Proof finished. $\hfill\qedsymbol$ 

\newpage
\section{Case Study}\label{Case_Study}
Here we show part of the case study. We use different colored fonts to enhance readability, where \textcolor{sys_prompt}{green fonts represent system prompts}, \textcolor{question}{purple fonts represent questions}, \textcolor{CoT}{orange fonts represent Chain of Thought}, and \textcolor{answer}{blue fonts represent answers}.
\begin{tcolorbox}[
    colframe=c1, 
    colback=blue!5!white, 
    coltitle=white, 
    fonttitle=\bfseries, 
    title=Case Study for base model \texttt{Qwen2.5-7B-Instruct-1M}, 
    arc=3mm, 
    boxrule=0.5mm 
]
\textbf{Question}: A very special island is inhabited only by knights and knaves. Knights always tell the truth, and knaves always lie. You meet 4 inhabitants: Henry, Jack, Amelia, and Evelyn. Henry was heard saying, "Jack is a knave if and only if Henry is a knight". Jack commented, "Henry is not a knight". Amelia expressed that Evelyn is a knave. In a statement by Evelyn: "Jack is a knight and Henry is a knight". So who is a knight and who is a knave?\\

\textbf{Answer}: (1) Henry is a knight, (2) Jack is a knave, (3) Amelia is a knight, (4) Evelyn is a knave. \\

\textcolor{sys_prompt}{<|im\_start|>system\textbackslash nYou are a helpful assistant.<|im\_end|>\textbackslash n<|im\_start|>user\textbackslash n<|im\_start|> system\textbackslash nYou are a helpful assistant. The assistant first thinks about the reasoning process in the mind and then provides the user with the answer. The reasoning process and answer are enclosed within <think> </think> and<answer> </answer> tags, respectively, i.e., <think> reasoning process here </think><answer> answer here </answer>.  Now the user asks you to solve a logical reasoning problem. After thinking, when you finally reach a conclusion, clearly state the identity of each character within <answer> </answer> tags. i.e., <answer> (1) Zoey is a knight\textbackslash n(2) ... </answer>.\textbackslash n<|im\_end|>\textbackslash n}\textcolor{question}{<|im\_start|>user\textbackslash nA very special island is inhabited only by knights and knaves. Knights always tell the truth, and knaves always lie. You meet 4 inhabitants: Henry, Jack, Amelia, and Evelyn. Henry was heard saying, "Jack is a knave if and only if Henry is a knight". Jack commented, "Henry is not a knight". Amelia expressed that Evelyn is a knave. In a statement by Evelyn: "Jack is a knight and Henry is a knight". So who is a knight and who is a knave?\textbackslash n<|im\_end|>\textbackslash n<|im\_start|>assistant\textbackslash n<think>}\textcolor{CoT}{<|im\_end|>\textbackslash n<|im\_start|>assistant\textbackslash nLet\'s analyze each statement one by one, considering the nature of knights (who always tell the truth) and knaves (who always lie).\textbackslash n\textbackslash n1. **Henry\'s statement:** "Jack is a knave if and only if Henry is a knight."\textbackslash n   - This is a conditional statement of the form \( A \iff B \) (A if and only if B). This means that both \( A \) and \( B \) must be true or both must be false.\textbackslash n   - If Henry is a knight, then Jack must also be a knight (because knights always tell the truth).\textbackslash n   - If Henry is a knave, then Jack must also be a knave (because knaves always lie, and the statement would be false).\textbackslash n\textbackslash n2. **Jack\'s statement:** "Henry is not a knight."\textbackslash n   - If Jack is a knight, this statement must be true, so Henry is indeed a knave.\textbackslash n   - If Jack is a knave, this statement must be false, so Henry is actually a knight.\textbackslash n\textbackslash n3. **Amelia\'s statement:** "Evelyn is a knave."\textbackslash n   - If Amelia is a knight, then Evelyn is indeed a knave.\textbackslash n   - If Amelia is a knave, then Evelyn is actually a knight.\textbackslash n\textbackslash n4. **Evelyn\'s statement:** "Jack is a knight and Henry is a knight."\textbackslash n   - If Evelyn is a knight, then both Jack and Henry must be knights.\textbackslash n   - If Evelyn is a knave, then either Jack or Henry (or both) must be a knave.\textbackslash n\textbackslash nLet\'s evaluate the scenarios:\textbackslash n\textbackslash n- If Henry is a knight:\textbackslash n  - According to Henry, Jack would also be a knight.\textbackslash n  - According to Evelyn, both Jack and Henry are knights. This would mean Evelyn is a knight.\textbackslash n  - According to Amelia, Evelyn is a knave, which contradicts the fact that Evelyn is a knight. Thus, Henry cannot be a knight.\textbackslash n\textbackslash n- If Henry is a knave:\textbackslash n  - According to Henry, Jack would also be a knave, which contradicts Jack\'s statement that Henry is not a knight. Thus, Jack cannot be a knave if Henry is a knave.\textbackslash n  - Jack must be a knight, as his statement "Henry is not a knight" would be true. \textbackslash n  - According to Evelyn, Jack and Henry are both knights, but we already established Henry is a knave. Thus, Evelyn must be a knave.\textbackslash n  - If Evelyn is a knave, her statement "Jack is a knight and Henry is a knight" is false, which is consistent with the fact that Jack is a knight and Henry is a knave.\textbackslash n  - This means Amelia must be a knight, as her statement "Evelyn is a knave" is true.\textbackslash n\textbackslash nSummarizing the results:\textbackslash n- Henry is a knave.\textbackslash n- Jack is a knight.\textbackslash n- Evelyn is a knave.\textbackslash n- Amelia is a knight.\textbackslash n\textbackslash n}\textcolor{answer}{<answer>\textbackslash n(1) Henry is a knave\textbackslash n(2) Jack is a knight\textbackslash n(3) Evelyn is a knave\textbackslash n(4) Amelia is a knight\textbackslash n</answer><|im\_end|>}
\end{tcolorbox}
\newpage

\begin{tcolorbox}[
    colframe=c1, 
    colback=blue!5!white, 
    coltitle=white, 
    fonttitle=\bfseries, 
    title=Case Study for model \texttt{Qwen2.5-7B-Instruct-1M+TRPA}, 
    arc=3mm, 
    boxrule=0.5mm 
]
\textbf{Question}: A very special island is inhabited only by knights and knaves. Knights always tell the truth, and knaves always lie. You meet 4 inhabitants: Henry, Jack, Amelia, and Evelyn. Henry was heard saying, "Jack is a knave if and only if Henry is a knight". Jack commented, "Henry is not a knight". Amelia expressed that Evelyn is a knave. In a statement by Evelyn: "Jack is a knight and Henry is a knight". So who is a knight and who is a knave?\\

\textbf{Answer}: (1) Henry is a knight, (2) Jack is a knave, (3) Amelia is a knight, (4) Evelyn is a knave. \\

\textcolor{sys_prompt}{<|im\_start|>system\textbackslash nYou are a helpful assistant.<|im\_end|>\textbackslash n<|im\_start|>user\textbackslash n<|im\_start|> system\textbackslash nYou are a helpful assistant. The assistant first thinks about the reasoning process in the mind and then provides the user with the answer. The reasoning process and answer are enclosed within <think> </think> and<answer> </answer> tags, respectively, i.e., <think> reasoning process here </think><answer> answer here </answer>.  Now the user asks you to solve a logical reasoning problem. After thinking, when you finally reach a conclusion, clearly state the identity of each character within <answer> </answer> tags. i.e., <answer> (1) Zoey is a knight\textbackslash n(2) ... </answer>.\textbackslash n<|im\_end|>\textbackslash n}\textcolor{question}{<|im\_start|>user\textbackslash nA very special island is inhabited only by knights and knaves. Knights always tell the truth, and knaves always lie. You meet 4 inhabitants: Henry, Jack, Amelia, and Evelyn. Henry was heard saying, "Jack is a knave if and only if Henry is a knight". Jack commented, "Henry is not a knight". Amelia expressed that Evelyn is a knave. In a statement by Evelyn: "Jack is a knight and Henry is a knight". So who is a knight and who is a knave?\textbackslash n<|im\_end|>\textbackslash n<|im\_start|>assistant\textbackslash n<think>}\textcolor{CoT}{<|im\_end|>\textbackslash n<|im\_start|>assistant\textbackslash nTo determine who is a knight and who is a knave, let's analyze the statements step by step.\textbackslash n\textbackslash n1. Henry says: "Jack is a knave if and only if Henry is a knight."\textbackslash nWe need to determine if Henry is a knight or knave.\textbackslash n\textbackslash nThere are two possibilities for Henry: he is a knight or he is a knave.\textbackslash n\textbackslash nCase 1: Henry is a knight.\textbackslash nThen, Jack is a knave (because the statement "Jack is a knave if and only if Henry is a knight" is true; Henry is a knight, so Jack is a knave).\textbackslash n\textbackslash nJack says: "Henry is not a knight."\textbackslash nWe determined that Henry is a knight. So, Jack is a knave.\textbackslash n\textbackslash nEvelyn says: "Jack is a knight and Henry is a knight."\textbackslash nWe determined that Jack is a knave. So, the first part (Jack is a knight) is false. Therefore, Evelyn is a knave.\textbackslash n\textbackslash nAmelia says: "Evelyn is a knave."\textbackslash nWe determined that Evelyn is a knave. So, Amelia is a knight. </think>}\textcolor{answer}{<answer> Henry is a knight, Jack is a knave, Amelia is a knight, Evelyn is a knave. </answer><|im\_end|>}
\end{tcolorbox}

\begin{tcolorbox}[
    colframe=c1, 
    colback=blue!5!white, 
    coltitle=white, 
    fonttitle=\bfseries, 
    title=Case Study for model \texttt{DeepSeek-R1-Distill-Qwen-7B(8k)+TRPA}, 
    arc=3mm, 
    boxrule=0.5mm 
]
\textbf{Question}: Every morning Aya goes for a $9$-kilometer-long walk and stops at a coffee shop afterwards. When she walks at a constant speed of $s$ kilometers per hour, the walk takes her 4 hours, including $t$ minutes spent in the coffee shop. When she walks $s+2$ kilometers per hour, the walk takes her 2 hours and 24 minutes, including $t$ minutes spent in the coffee shop. Suppose Aya walks at $s+\frac{1}{2}$ kilometers per hour. Find the number of minutes the walk takes her, including the $t$ minutes spent in the coffee shop.\\

\textbf{Answer}: 204 \\

\textcolor{question}{<|begin\_of\_sentence|><|User|>Every morning Aya goes for a $9$-kilometer-long walk and stops at a coffee shop afterwards. When she walks at a constant speed of $s$ kilometers per hour, the walk takes her 4 hours, including $t$ minutes spent in the coffee shop. When she walks $s+2$ kilometers per hour, the walk takes her 2 hours and 24 minutes, including $t$ minutes spent in the coffee shop. Suppose Aya walks at $s+\frac{1}{2}$ kilometers per hour. Find the number of minutes the walk takes her, including the $t$ minutes spent in the coffee shop. }\textcolor{sys_prompt}{Let's think step by step and output the final answer within \textbackslash\textbackslash boxed\{\}.}\textcolor{CoT}{<|Assistant|><think>\textbackslash nOkay, so Aya walks 9 kilometers every morning and then stops at a coffee shop. She walks at speed s km/h, and the whole trip, including the coffee shop time, takes 4 hours. When she walks faster, s+2 km/h, the total time is 2 hours and 24 minutes, which is 2.4 hours, I think. Wait, 24 minutes is 0.4 hours, so total time is 2.4 hours? Or wait, 2 hours 24 minutes is 2 + 24/60 = 2.4 hours, yes.\textbackslash n\textbackslash nSo, the time she spends walking plus the coffee shop time t is the total time. So, for the first case: walking time is 9/s hours, plus t minutes. But wait, the total time is 4 hours, which is 4*60=240 minutes. So, 9/s hours is (9/s)*60 minutes. So, (9/s)*60 + t = 240.\textbackslash n\textbackslash nIn the second case, walking speed is s+2, so walking time is 9/(s+2) hours, which is 9*60/(s+2) minutes. Total time is 2 hours 24 minutes, which is 144 minutes (since 2*60 +24=144). So, 9*60/(s+2) + t = 144.\textbackslash n\textbackslash nNow, we have two equations:\textbackslash n\textbackslash n1) (540)/s + t = 240\textbackslash n2) (540)/(s+2) + t = 144\textbackslash n\textbackslash nSubtracting equation 2 from equation 1:\textbackslash n\textbackslash n(540/s) - (540/(s+2)) = 240 - 144 = 96\textbackslash n\textbackslash nSo, 540*(1/s - 1/(s+2)) = 96\textbackslash n\textbackslash nSimplify 1/s - 1/(s+2) = (2)/(s(s+2)).\textbackslash n\textbackslash nSo, 540*(2)/(s(s+2)) = 96\textbackslash n\textbackslash nSo, $1080 / (s^2 + 2s) = 96$\textbackslash n\textbackslash nMultiply both sides by denominator:\textbackslash n\textbackslash n$1080 = 96(s^2 + 2s)$\textbackslash n\textbackslash nDivide both sides by 24: $45 = 4(s^2 + 2s)$\textbackslash n\textbackslash nSo, $4s^2 + 8s -45 =0$\textbackslash n\textbackslash nWait, wait: 96/24 is 4, so 1080/24 is 45. So, $45 = 4s^2 + 8s$.\textbackslash n\textbackslash nSo, $4s^2 +8s -45=0$.\textbackslash n\textbackslash nSolving quadratic equation: s = [-8 ± sqrt(64 + 720)]/(2*4) = [-8 ± sqrt(784)]/8 = [-8 ±28]/8.\textbackslash n\textbackslash nSince speed can't be negative, so positive solution: (20)/8 = 2.5 km/h.\textbackslash n\textbackslash nSo, s=2.5 km/h.\textbackslash n\textbackslash nNow, find t. From equation 1: (540)/2.5 + t = 240.\textbackslash n\textbackslash n540/2.5 is 540*2/5= 216.\textbackslash n\textbackslash nSo, 216 + t = 240 => t=24 minutes.\textbackslash n\textbackslash nNow, when she walks at s + 0.5 km/h, which is 2.5 + 0.5=3 km/h.\textbackslash n\textbackslash nWalking time: 9/3 =3 hours, which is 180 minutes. Plus t=24, so total time 204 minutes.\textbackslash n</think>\textbackslash n\textbackslash n}\textcolor{answer}{\textbackslash\textbackslash boxed\{204\}}
\end{tcolorbox}

\begin{tcolorbox}[
    colframe=c1, 
    colback=blue!5!white, 
    coltitle=white, 
    fonttitle=\bfseries, 
    title=Case Study for model \texttt{DeepSeek-R1-Distill-Qwen-7B(32k)}, 
    arc=3mm, 
    boxrule=0.5mm 
]
\textbf{Question}: Every morning Aya goes for a $9$-kilometer-long walk and stops at a coffee shop afterwards. When she walks at a constant speed of $s$ kilometers per hour, the walk takes her 4 hours, including $t$ minutes spent in the coffee shop. When she walks $s+2$ kilometers per hour, the walk takes her 2 hours and 24 minutes, including $t$ minutes spent in the coffee shop. Suppose Aya walks at $s+\frac{1}{2}$ kilometers per hour. Find the number of minutes the walk takes her, including the $t$ minutes spent in the coffee shop.\\

\textbf{Answer}: 204 \\

\textcolor{question}{<|begin\_of\_sentence|><|User|>Every morning Aya goes for a $9$-kilometer-long walk and stops at a coffee shop afterwards. When she walks at a constant speed of $s$ kilometers per hour, the walk takes her 4 hours, including $t$ minutes spent in the coffee shop. When she walks $s+2$ kilometers per hour, the walk takes her 2 hours and 24 minutes, including $t$ minutes spent in the coffee shop. Suppose Aya walks at $s+\frac{1}{2}$ kilometers per hour. Find the number of minutes the walk takes her, including the $t$ minutes spent in the coffee shop. }\textcolor{sys_prompt}{Let's think step by step and output the final answer within \textbackslash\textbackslash boxed\{\}.}\textcolor{CoT}{<|Assistant|><think>\textbackslash nOkay, let's try to figure this problem out step by step. So, Aya goes for a 9-kilometer walk every morning and then stops at a coffee shop. Her walking speed affects the total time, including the time spent in the coffee shop.\textbackslash n\textbackslash nThe problem gives two scenarios. In the first, she walks at s km/h, and the total time is 4 hours, which includes t minutes at the coffee shop. In the second scenario, she walks at (s + 2) km/h, and the total time is 2 hours and 24 minutes, including the same t minutes in the coffee shop.\textbackslash n\textbackslash nWe need to find the total time when she walks at (s + 1/2) km/h, again including t minutes. So, we need to find s, t, and then compute the time for the third scenario.\textbackslash n\textbackslash nFirst, let's denote:\textbackslash n\textbackslash n- Walking distance: 9 km.\textbackslash n- Time when walking at s km/h: total time is 4 hours, which is 4*60=240 minutes.\textbackslash n- Time when walking at (s+2) km/h: total time is 2h24m, which is 2*60 +24=144 minutes.\textbackslash n\textbackslash nSince both include t minutes, the walking time is total time - t.\textbackslash n\textbackslash nSo, for the first case:\textbackslash nWalking time = 240 - t minutes. But since speed is in km/h, we need to convert walking time to hours.\textbackslash n\textbackslash nSo, 240 - t (in minutes) = 9 / s hours.\textbackslash n\textbackslash nSimilarly, for the second case:\textbackslash nWalking time = 144 - t (in minutes) = 9 / (s + 2) hours.\textbackslash n\textbackslash nWait, but the units here are a bit mixed. Let me correct that. Since 240 - t is in minutes, we need to convert it to hours by dividing by 60.\textbackslash n\textbackslash nSo, equation 1: (240 - t)/60 = 9 / s\textbackslash n\textbackslash nEquation 2: (144 - t)/60 = 9 / (s + 2)\textbackslash n\textbackslash nNow, we have two equations with two unknowns, s and t. Let's write them as:\textbackslash n\textbackslash nEquation 1: (240 - t)/60 = 9/s => 240 - t = (540)/s\textbackslash n\textbackslash nEquation 2: (144 - t)/60 = 9/(s + 2) => 144 - t = (540)/(s + 2)\textbackslash n\textbackslash nSo, from equation 1: 240 - t = 540/s --> t = 240 - (540/s)\textbackslash n\textbackslash nFrom equation 2: 144 - t = 540/(s+2) --> t = 144 - 540/(s+2)\textbackslash n\textbackslash nSince both expressions equal t, we can set them equal:\textbackslash n\textbackslash n240 - 540/s = 144 - 540/(s + 2)\textbackslash n\textbackslash nLet's solve for s.\textbackslash n\textbackslash n240 - 144 = 540/s - 540/(s + 2)\textbackslash n\textbackslash n96 = 540 [1/s - 1/(s + 2)]\textbackslash n\textbackslash nFactor 540: 540 = 54*10=540.\textbackslash n\textbackslash nSo,\textbackslash n\textbackslash n96 = 540 [ (s + 2 - s) / (s(s +2)) ) ]\textbackslash n\textbackslash nSimplify numerator: 2.\textbackslash n\textbackslash nSo,\textbackslash n\textbackslash n96 = 540 * (2) / (s(s +2))\textbackslash n\textbackslash n$96 = 1080 / (s^2 + 2s)$\textbackslash n\textbackslash nMultiply both sides by ($s^2 + 2s$):\textbackslash n\textbackslash n96($s^2 + 2s$) = 1080\textbackslash n\textbackslash nDivide both sides by 96:\textbackslash n\textbackslash n$s^2 + 2s = 1080 / 96$\textbackslash n\textbackslash nSimplify 1080/96: divide numerator and denominator by 24: 45/4.\textbackslash n\textbackslash nSo,\textbackslash n\textbackslash n$s^2 + 2s = 45/4$\textbackslash n\textbackslash nMultiply both sides by 4:\textbackslash n\textbackslash n$4s^2 + 8s = 45$\textbackslash n\textbackslash nBring all to left:\textbackslash n\textbackslash n$4s^2 +8s -45 =0$\textbackslash n\textbackslash nQuadratic equation: $4s^2 +8s -45=0$\textbackslash n\textbackslash nSolve for s.\textbackslash n\textbackslash nUsing quadratic formula: $s = [-b ± sqrt(b^2 -4ac)]/(2a)$\textbackslash n\textbackslash na=4, b=8, c=-45.\textbackslash n\textbackslash nDiscriminant: 64 -4*4*(-45) = 64 + 720=784\textbackslash n\textbackslash nsqrt(784)=28\textbackslash n\textbackslash nSo,\textbackslash n\textbackslash ns = [ -8 ±28 ]/(8)\textbackslash n\textbackslash nWe can ignore the negative solution since speed can't be negative.\textbackslash n\textbackslash nSo, s = (20)/8 = 2.5 km/h.\textbackslash n\textbackslash nSo, s=2.5 km/h.\textbackslash n\textbackslash nNow, find t.\textbackslash n\textbackslash nFrom equation 1: t = 240 - 540/s\textbackslash n\textbackslash ns=2.5, so 540/2.5=216.\textbackslash n\textbackslash nt=240-216=24 minutes.\textbackslash n\textbackslash nSo, t=24.\textbackslash n\textbackslash nNow, when she walks at s + 1/2 = 2.5 +0.5=3 km/h.\textbackslash n\textbackslash nWalking time: 9/3=3 hours, which is 180 minutes.\textbackslash n\textbackslash nTotal time including t: 180 +24=204 minutes.\textbackslash n\textbackslash nSo, the answer is 204 minutes.\textbackslash n\textbackslash nWait, let me check.\textbackslash n\textbackslash nWait, when she walks at 2.5 km/h, the walking time is 9/2.5=3.6 hours=216 minutes. Then coffee shop time is t=24, so total 240, which matches.\textbackslash n\textbackslash nWhen she walks at 4.5 km/h (wait, s+2=2.5+2=4.5), 9/4.5=2 hours=120 minutes, plus 24 minutes: total 144, which is correct.\textbackslash n\textbackslash nSo, the third case: 3 km/h.\textbackslash n\textbackslash nWalking time: 9/3=3 hours=180 minutes.\textbackslash n\textbackslash nPlus t=24: total 204 minutes. So, 204 is the answer.\textbackslash n</think>\textbackslash n\textbackslash nThe walk takes 204 minutes.}\textcolor{answer}{\textbackslash\textbackslash boxed\{204\}}
\end{tcolorbox}

\begin{tcolorbox}[
    colframe=c1, 
    colback=blue!5!white, 
    coltitle=white, 
    fonttitle=\bfseries, 
    title=Case Study for model \texttt{DeepSeek-R1-Distill-Qwen-7B(8k)+TRPA}, 
    arc=3mm, 
    boxrule=0.5mm 
]
\textbf{Question}: Alice and Bob play the following game. A stack of $n$ tokens lies before them. The players take turns with Alice going first. On each turn, the player removes either $1$ token or $4$ tokens from the stack. Whoever removes the last token wins. Find the number of positive integers $n$ less than or equal to $2024$ for which there exists a strategy for Bob that guarantees that Bob will win the game regardless of Alice's play.\\

\textbf{Answer}: 809\\

\textcolor{question}{<|begin\_of\_sentence|><|User|>Alice and Bob play the following game. A stack of $n$ tokens lies before them. The players take turns with Alice going first. On each turn, the player removes either $1$ token or $4$ tokens from the stack. Whoever removes the last token wins. Find the number of positive integers $n$ less than or equal to $2024$ for which there exists a strategy for Bob that guarantees that Bob will win the game regardless of Alice's play.}\textcolor{sys_prompt}{Let's think step by step and output the final answer within \textbackslash\textbackslash boxed\{\}.}\textcolor{CoT}{ <|Assistant|><think>\textbackslash nOkay, so Alice and Bob are playing a game where they take turns removing either 1 or 4 tokens from a stack of n tokens. The person who takes the last token wins. We need to find how many positive integers n $\leq$ 2024 where Bob can force a win, no matter how Alice plays.\textbackslash n\textbackslash nHmm, let's think about this. It seems like a classic game theory problem, specifically a variant of the take-away game. The key here is to determine for each n whether it's a winning or losing position.\textbackslash n\textbackslash nA position is losing if every possible move leads to a winning position for the opponent. Conversely, a position is winning if there's at least one move that leads the opponent into a losing position.\textbackslash n\textbackslash nLet's denote the positions as losing (L) or winning (W). We need to find the pattern.\textbackslash n\textbackslash nLet's start with small n:\textbackslash n\textbackslash nn=1: Alice takes 1 and wins. So, n=1 is a winning position (W).\textbackslash n\textbackslash nn=2: Alice can take 1, leaving 1 for Bob, which is W for Bob. Or take 4, which isn't possible. Wait, wait, no. Wait, n=2: Alice can only take 1, leaving 1. Then Bob takes 1 and wins. So for n=2, Alice's move leads to Bob winning. Wait, wait, no. Wait, n=2: if Alice takes 1, n becomes 1, Bob takes 1 and wins. So n=2: Alice can't force a win, so n=2 is a losing position? Wait, no, because if it's Alice's turn, she can only take 1, leaving 1, which is a W position for Bob. So n=2 is a losing position for Alice, meaning Bob can win. So n=2 is L.\textbackslash n\textbackslash nWait, but I'm getting confused. Let me clarify: in this problem, Alice goes first. So for each n, if the starting player can force a win, it's W; else, L.\textbackslash n\textbackslash nSo let's redo:\textbackslash n\textbackslash nn=1: Alice takes 1 and wins. So W.\textbackslash n\textbackslash nn=2: Alice takes 1, leaves 1. Bob takes 1 and wins. So for n=2, Alice can't win. So n=2 is L.\textbackslash n\textbackslash nn=3: Alice can take 1, leaving 2 (which is L for Bob, because Bob is now facing n=2, which is a losing position. So if Alice takes 1, Bob is in a losing position, so Alice can win. So n=3: W.\textbackslash n\textbackslash nn=4: Alice can take 4 and win. So W.\textbackslash n\textbackslash nn=5: Let's see. Alice can take 1, leaving 4, which is W for Bob. Or take 4, leaving 1, which is W for Bob. So no matter what, Bob can win. So n=5 is L.\textbackslash n\textbackslash nn=6: Alice can take 1, leaving 5 (L for Bob). So Bob is in a losing position. So Alice can win. So n=6: W.\textbackslash n\textbackslash nn=7: Alice can take 1, leaving 6 (W for Bob). Or take 4, leaving 3 (W for Bob). So Bob can win in either case. So n=7: L.\textbackslash n\textbackslash nn=8: Alice can take 1, leaving 7 (L for Bob). So Bob is in losing position. So n=8: W.\textbackslash n\textbackslash nn=9: Alice can take 1, leaving 8 (W for Bob). Or take 4, leaving 5 (L for Bob). So Alice can choose to take 4, leaving Bob with n=5, which is L. So n=9: W.\textbackslash n\textbackslash nn=10: Alice can take 1, leaving 9 (W for Bob). Or take 4, leaving 6 (W for Bob). So Bob can win. So n=10: L.\textbackslash n\textbackslash nHmm, I'm seeing a pattern here. Let's list the positions:\textbackslash n\textbackslash nn | outcome (W or L)\textbackslash n1 | W\textbackslash n2 | L\textbackslash n3 | W\textbackslash n4 | W\textbackslash n5 | L\textbackslash n6 | W\textbackslash n7 | L\textbackslash n8 | W\textbackslash n9 | W\textbackslash n10| L\textbackslash n\textbackslash nLooking at this, it seems like every 5 numbers, the pattern repeats. Let me check:\textbackslash n\textbackslash nFrom n=1 to 5: W, L, W, W, L.\textbackslash n\textbackslash nn=6: W, n=7: L, n=8: W, n=9: W, n=10: L.\textbackslash n\textbackslash nYes, the pattern repeats every 5 numbers. So the cycle is W, L, W, W, L, W, L, W, W, L, etc.\textbackslash n\textbackslash n}
\end{tcolorbox}
\begin{tcolorbox}[
    colframe=c1, 
    colback=blue!5!white, 
    coltitle=white, 
    fonttitle=\bfseries, 
    title=Continued from the above box, 
    arc=3mm, 
    boxrule=0.5mm 
]
\textcolor{CoT}{So the losing positions are n=2,5,7,10,12,15,... Wait, wait, let's see:\textbackslash n\textbackslash nWait, n=2: L\textbackslash n\textbackslash nn=5: L\textbackslash n\textbackslash nn=7: L\textbackslash n\textbackslash nn=10: L\textbackslash n\textbackslash nWait, the differences between these are 3, 2, 3, etc. Hmm, maybe the pattern is that every 5 steps, the losing positions are at n = 2 or 0 mod 5? Wait, let's see:\textbackslash n\textbackslash nn=2: 2 mod 5 is 2\textbackslash n\textbackslash nn=5: 0 mod5\textbackslash n\textbackslash nn=7: 2 mod5\textbackslash n\textbackslash nn=10: 0 mod5\textbackslash n\textbackslash nn=12: 2 mod5?\textbackslash n\textbackslash nYes, so the losing positions are n = 0 or 2 mod5.\textbackslash n\textbackslash nWait, 5 mod5 is 0, 7 mod5 is 2, 10 mod5 is 0, 12 mod5 is 2, etc.\textbackslash n\textbackslash nSo the losing positions are n = 0 or 2 mod5.\textbackslash n\textbackslash nTherefore, the number of losing positions up to 2024 is the count of n $\leq$2024 where n mod5 is 0 or 2.\textbackslash n\textbackslash nSo how many numbers $\leq$2024 are =0 or 2 mod5.\textbackslash n\textbackslash nFirst, for 0 mod5: numbers divisible by5. The count is floor(2024/5) = 404 (since 5*404=2020, which is $\leq$2024).\textbackslash n\textbackslash nFor 2 mod5: numbers like 2,7,12,... up to $\leq$2024.\textbackslash n\textbackslash nThe maximum number is the largest number $\leq$2024 which is 2 mod5.\textbackslash n\textbackslash n2024 divided by5 is 404.8, so 5*404=2020, 2020+2=2022. So 2022 is the last such number.\textbackslash n\textbackslash nThe count is same as the number of terms in 2,7,12,...,2022. The number of terms is (2022 -2)/5 +1 = (2020)/5 +1 = 404 +1=405.\textbackslash n\textbackslash nSo total losing positions: 404 (for 0 mod5) +405 (for 2 mod5) = 809.\textbackslash n\textbackslash nWait, 404 +405=809.\textbackslash n\textbackslash nWait, but 2024=5*404 +4, so 2024 is 4 mod5, which is not 0 or 2. So the counts are correct.\textbackslash n\textbackslash nTherefore, the number of n $\leq$2024 where Bob can win is 809.\textbackslash n\textbackslash nSo the answer is 809.\textbackslash n</think>\textbackslash n\textbackslash n}\textcolor{answer}{The number of positive integers n $\leq$ 2024 where Bob can force a win is \textbackslash\textbackslash boxed\{809\}.}
\end{tcolorbox}



\end{document}